\documentclass[a4paper,fleqn]{cas-dc}

\usepackage[numbers]{natbib}

\def\tsc#1{\csdef{#1}{\textsc{\lowercase{#1}}\xspace}}
\tsc{WGM}
\tsc{QE}
\tsc{EP}
\tsc{PMS}
\tsc{BEC}
\tsc{DE}

\begin{document}
\let\WriteBookmarks\relax
\def\floatpagepagefraction{1}
\def\textpagefraction{.001}
\shorttitle{3D Human Motion Prediction : A Survey}
\shortauthors{Kedi Lyu et~al.}

\title [mode = title]{3D Human Motion Prediction : A Survey}      



\author[1]{Kedi Lyu}[type=editor,
                        auid=000,bioid=1,
                        orcid=0000-0002-7905-3759
                        ]
\fnmark[1]
\ead{lvkd19@mails.jlu.edu.cn}

\credit{Conceptualization of this study, Methodology, Software}

\address[1]{Jilin University, Changchun, Jilin, China}

\author[1]{Haipeng Chen}

   

\ead{chenhp@jlu.edu.cn}

\credit{Data curation, Writing - Original draft preparation}

\address[2]{Zhejiang Gongshang University, Hangzhou, Zhejiang, China}

\address[3]{Sichuan University, Chengdu, Sichuan, China}

\author[2]
{Zhenguang Liu}
\cormark[1]
\ead{liuzhenguang2008@gmail.com}

\author[3]
{Beiqi Zhang}
\ead{beiqizhang126@126.com}

\author[2]
{Ruili Wang}
\cormark[1]
\ead{ruili.wang@massey.ac.nz}


\cortext[cor1]{Corresponding author}
\fntext[fn1]{This is the first author footnote. }


\begin{abstract}
3D human motion prediction, predicting future poses from a given sequence, is an issue of great significance and challenge in computer vision and machine intelligence, which can help machines in understanding human behaviors. Due to the increasing development and understanding of Deep Neural Networks (DNNs) and the availability of large-scale human motion datasets, the human motion prediction has been remarkably advanced with a surge of interest among academia and industrial community.
In this context, a comprehensive survey on 3D human motion prediction is conducted for the purpose of retrospecting and analyzing relevant works from existing released literature. In addition, a pertinent taxonomy is constructed to categorize these existing approaches for 3D human motion prediction. 
In this survey, relevant methods are categorized into  three categories: \textit{human pose representation}, \textit{network structure design}, and \textit{prediction target}. 
We systematically review all relevant journal and conference papers in the field of human motion prediction since 2015, which are presented in detail based on proposed categorizations in this survey.
Furthermore, the outline for the public benchmark datasets, evaluation criteria, and performance comparisons are respectively presented in this paper. The limitations of the state-of-the-art methods are discussed as well, hoping for paving the way for future explorations.  
\end{abstract}

\maketitle
\section{Introduction}
\begin{figure}
	\centering
		\includegraphics[scale=0.34]{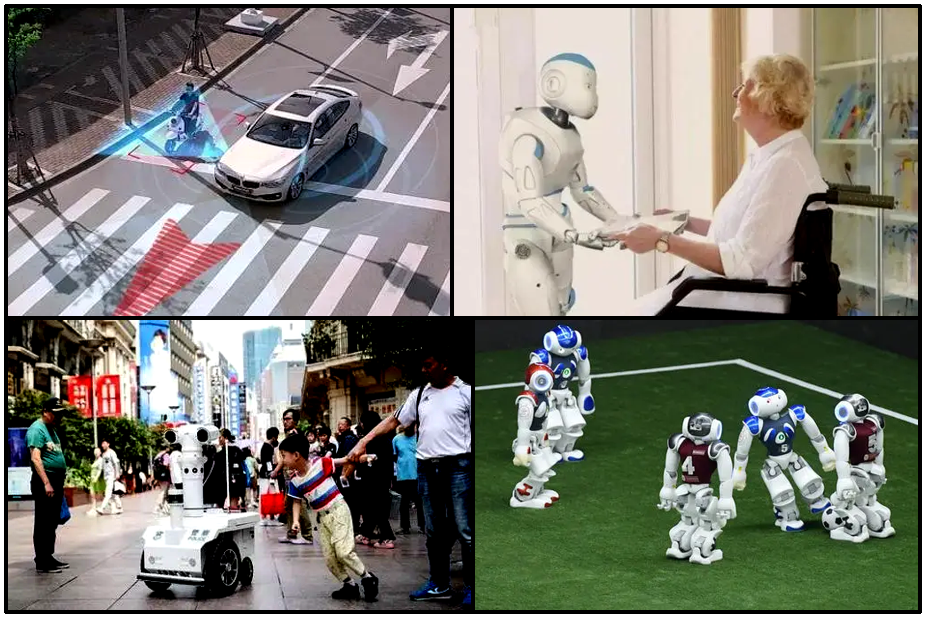}
	\caption{Applications of 3D human motion prediction. 
	\textbf{Top left}: Autonomous vehicles need to promptly judge the intentions and future positions of other traffic participants. 
	\textbf{Top right}: Robots assisting humans, such as delivering documents to a human, need to predict human motion and accurately place documents in a given location.  
	\textbf{Bottom left}: In densely populated spaces, machines should accurately predict the human motions around them to safely pass through the crowds. \textbf{Bottom right}: In the RoboCup, robots need to predict the actions of opponents to perform effective offensive, and defensive behaviors.}
	\label{FIG:app}
\end{figure}

Comprehending and predicting human motion is crucial in assisting humans and machines to interact with the outside world. Predicting future human motions is an innate ability for humans to interact with other people such as navigating through the crowds, defending against offensive players in a game, or shaking hands with others. Further, it is also important for intelligent machines to respond to human behaviors, coordinate their poses, and project paths during the interactions with humans. However, as a non-talent ability of the machine, applying this ability is a very challenging task.
Further more, motion prediction has also been widely applied to autonomous driving \cite{driv1, driv2,DR3}, intelligent robot \cite{rob1, rob2}, human-robot collaboration \cite{HR1, HR2, HR3, HR4, HR5, HR6, HR7, HR8}, and multimedia applications \cite{DMT1,DMT2}, as shown in Fig \ref{FIG:app}.

The research of motion prediction has  attracted increasing attentions from academia and industry. Due to the advances in deep learning, great progress has been made in motion prediction. However,  it is still challenging to predict motions accurately. For instance, humans' intentions are very complex, which act as internal stimuli to drive human to behave differently. Likewise, the surroundings of the physical world can also affect humans' motions actively or passively. In fact, various factors that affect human poses  cannot always be intuitively identified or resorted by modeling the context. These indeed increase the intricacy of the issue, but provides diverse perspectives for our investigation. 

\subsection{Scope}
There are many  ways to interpret and represent human motions, such as kinematic trees \cite{AMGAN, SKEL}, joints graph \cite{ST-gcn, liu2021iccv}, video frames \cite{VHM1, VHM2}, and moving as a mass point from a starting point to a target point, which all reflect people's different comprehensions of human motion, resulting in diverse manifestations of human motion prediction.  
These have been inspired by various classifications of prediction methods: 
1) 2D Motion trajectory prediction: it mainly predicts  motion trajectories for human or moving devices \cite{2D1,2D2,2D3,2D4,2D5,2D6,2D7} in a 2D plane. 
2) Video prediction: it focuses on the motion prediction on the video frames \cite{vh1,vh2,vh3,vh4}. 
3) Poses sequence prediction: it predicts future 3D human motions. 
3D human motion prediction is the only scope in this survey.
The other two will not be included into the discussion scope and the conventional non-deep learning methods will not be taken into consideration here. As 3D human motion prediction is an emerging research field,  this survey will embrace as much literature as possible to provide a better understanding of current work and offer reference for further research in this field.  

\begin{figure}
	\centering
		\includegraphics[scale=0.5]{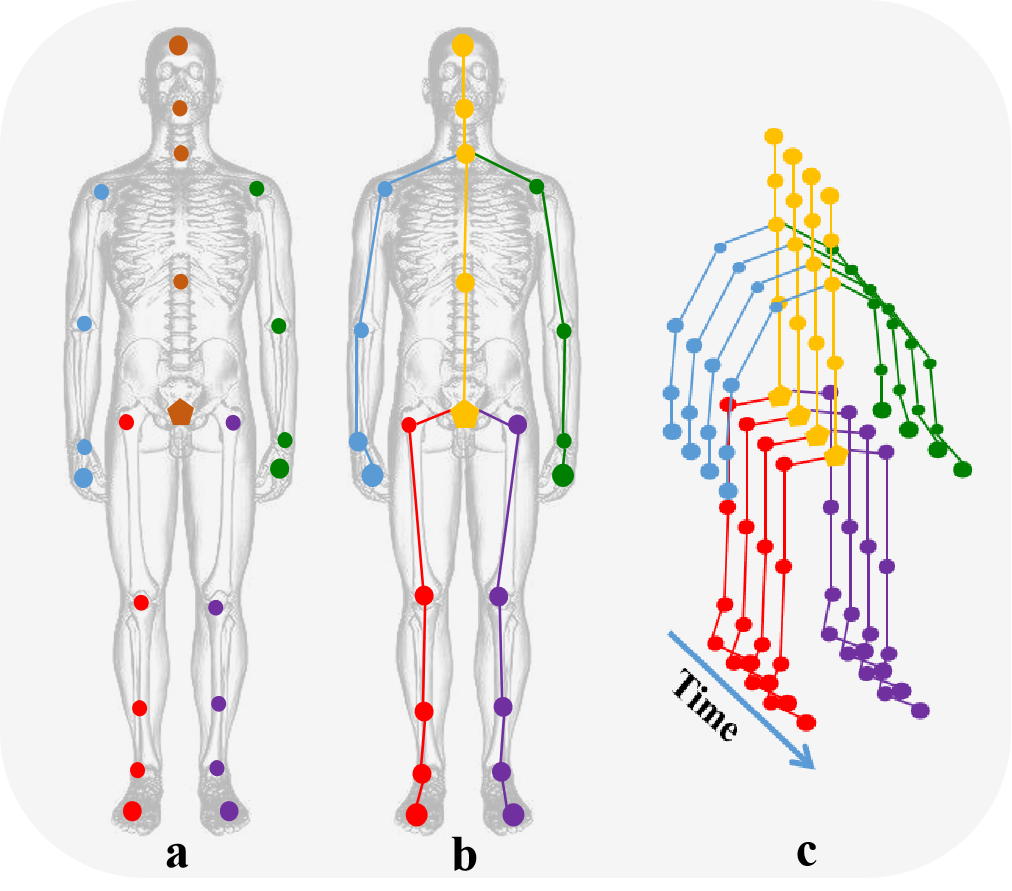}
	\caption{As is shown in the human body structure \textbf{a} and \textbf{b}, the human body is composed of 3D joints and different colored joints which constitute different kinematic chains, such as the central torso, the left/right arms and the left/right legs. The human body contains different numbers of joints in different datasets. For \textbf{c}, it is the human motion sequence made up of 3D human bodies that changes over time.}
	\label{FIG:body}
\end{figure}

\subsection{Problem formulation and challenges}
In 3D human motion prediction, it is a traditional method to represent the human pose by a skeletal kinematic tree composed of joints. 
In anatomy, the human body contains hundreds of joints, but only the joints recorded by a motion capture system will be focused on in this study. As shown in Fig \ref{FIG:body}, the human skeleton consists of representative joints. 
3D human motion is a sequence constituted by multiple human skeletons arranged in a chronological order. Mathematically, we suppose that a pose sequence of length $N$ is formulated as $X_{1:N} = (x_1,...,x_N)\in R^{J\times N\times D}$ and each $x_i \in R^{J\times D}$ belonging to sequence $X$, where $x_i$ represents a pose at the $i^{th}$ frame; $J$ is the number of joints; $D$ is the dimension number of each joint; $i\in (1,N)$. The future pose sequence is $X_{1+K:N+K} = (x_{K+1},...,x_{K+N})\in R^{J\times K\times D}$, where $K$ denotes the frames of the future sequence. The objective of 3D human motion prediction is to generate future sequences on the basis of the already observed ones. Mathematically, motion prediction aims to construct a model $F$ to forecast $\hat{X}_{1+K:N+K} = F(X_{1:N})$, where $\hat{X}_{1+K:N+K}$ is the predicted future motion sequence close to the ground truth $X_{1+K:N+K}$. 

The challenge of obtaining accurate and natural sequences of future human poses lies in the complexity of human behavior and the flexibility of human body.
As a result, understanding and predicting human motion is not an easy task for both humans and machines. 
The two key issues that need to be addressed are \textit{Inherent kinematic problems} and \textit{Network performance limitations}. 
The inherent kinematic problems of human motion prediction are triggered by its highly stochastic nature, high dimension, and non-linearity, which leads to a high degree of uncertainty for future human poses. 
Network performance limitations result from the inherent networks drawbacks that inevitably involve error accumulation from RNNs \cite{oRNN} and are stuck due to the limitation in processing standard 2D grid representations from CNNs. 

To cope with the challenges mentioned above, different solutions have been proposed. Initially, researchers focused on modeling human motion sequence-to-sequence, which depends on a deep RNN-based architecture that specializes in sequential problems.
On account of the prosperity of generative adversarial networks (GANs) \cite{oGan}, they are employed as a new kind of learning algorithm for human motion prediction.
Further, considering the connections between joints, Graph convolution networks (GCNs) \cite{ST-gcn}, a generalization of CNNs, are utilized to model the correlation between joints for the human motion prediction. Meanwhile, other CNNs \cite{Cs2s, CHA} methods are also proposed.

\subsection{Outline}
The purpose of this paper is to survey the existing methods of 3D human motion prediction and investigate these methods by classifying them and analyzing their performance differences. Then, the public benchmark datasets and evaluation metrics in this field are also reviewed in detail. This review intends to conduce to an accessible comprehension about the similarities and differences among a variety of methods and provides more ideas for future research.

The main contributions of this paper are summarized as follows: 

1) To the best of our knowledge, this paper should be hitherto the first survey on 3D human motion prediction. 

2) Existing methods in this domain are classified into three categories, and the benchmark datasets and evaluation metrics available in this domain are elaborated.  

3) Comparisons of existing methods are presented, and future research directions are discussed.

The remaining parts of the paper are organized as follows: 
The taxonomy of 3D human motion prediction is introduced in Section 2. 
The review and analysis of existing prediction methods are described minutely in Section 3. 
The experimental benchmark datasets and evaluation metrics are presented in Section 4.
Then, the comparison of different human motion prediction methods is shown in Section 5. 
In Section 6, the most advanced methods are discussed and the possible challenges for further research are listed. Finally, Section 7 is the conclusion of this paper.

\section{Taxonomy}
\begin{figure*}
	\centering
	\includegraphics[scale=0.75]{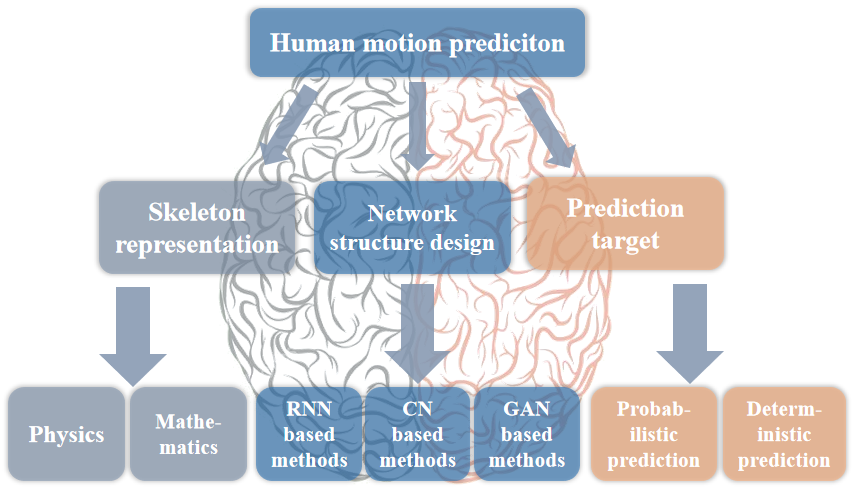}
	\caption{Overview of the categories in our taxonomy.}
	\label{FIG:methods}
\end{figure*}

In this section, the proposed taxonomy classifies human motion prediction methods into three categories: \textit{Human pose representations}, \textit{Network structure design}, and \textit{Prediction targets}, as shown in Fig.\ref{FIG:methods}. Afterwards, the explanation for the rules of taxonomy are introduced. Finally, the three categories are utilized as hints to further expound on the correlative methods, and representative papers in each category are taken as examples to promote understanding. 

\begin{figure*}
	\centering
	\includegraphics[scale=0.64]{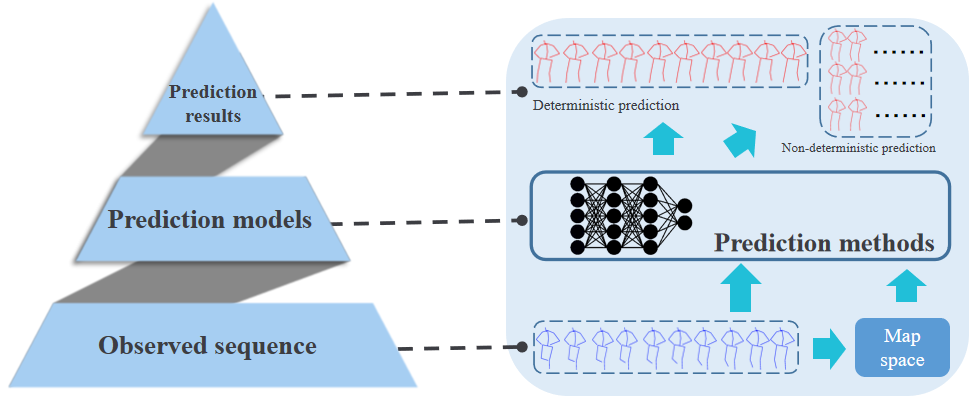}
	\caption{Generalization of the 3D human motion prediction task. On the \textbf{left}, it is a concise summary of the task. On the \textbf{right}, it is an overview of operations of methods.}
	\label{FIG:class}
\end{figure*}

\subsection{Classification rules}
\begin{figure}
		\includegraphics[scale=.75]{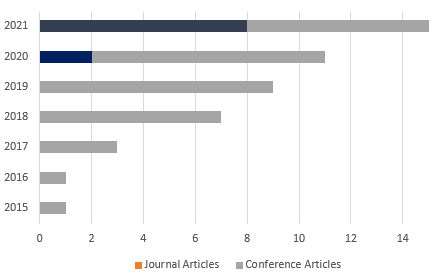}
	\caption{Publications trends in the literature reviewed for this survey.}
	\label{FIG:3}
\end{figure}

In recent years, as shown in Fig \ref{FIG:3}, human motion prediction has attracted more and more researchers to tackle this task with various approaches, which are generally a mixture of numerous different and basic ways. Hence, certain criteria to systematically classify them is urgently needed. We use the following rules:

1) Priority is given to modeling approaches. The approaches from the category introducing the modeling approaches are classified with precedence.

2) Focus is on the major models in the composite-model approach. Considering that some methods are combined with sub-components, the major models and innovations should be the focus for classification.

3) Emphasis is given to the diversity of perceptions. Different perceptions of prediction targets and human body representations determine the approaches to problem solving, which motivates our classification.  

With the highest level of abstraction, as shown in Fig.\ref{FIG:class}, the 3D human motion prediction task can be classified into three phases: \textit{Observed sequences}, \textit{Prediction models}, and \textit{Prediction results}. Three main problem-solving strategies are inspired among current research results, which are summarized in this paper as: \textit{Human pose representations}, \textit{Network structure design}, and \textit{Prediction targets}. 
Human pose representations emerge as the physical or mathematical transformation of parameterized observed human pose sequences. 
The network structure design aims at extracting and processing powerful features to facilitate prediction. It is the core means for the prediction models.  
The prediction target is not solitary. For a given past, there will be one or more generating sequences of future frames.

\subsection{Human pose representation}
All the while, human pose representation plays a fundamental yet important role in the field of computer vision.
An efficient human pose representation is helpful for the machine to understand the human behaviours. 
After induction, existing human pose representation approaches are classified into two categories: \textit{physical representation} and \textit{mathematical representation}.

\textbf{Physical representation} \quad
As a classic, skeleton-based human pose representations have been widely recognized. These benefit from their robustness to position display, kinematic representation, and real-time performance in the field of human motion prediction.     
Physical representations of human pose reflect the kinematic laws, basic structures, and motion forms. Employing natural 3D joint positions to denote the human pose skeleton is one of the most commonly used strategies. 
As a follow up of this line of work, hierarchical body parts are selected. 
Besides, being important motion properties, velocity and acceleration carry significant connotations of dynamics. 
As is shown in Tab \ref{tb math}, representative papers in this category are: SkelNet \cite{SKEL}, CHA \cite{CHA}, MGCN \cite{MGCN}, AM-GAN \cite{AMGAN}.  

\textbf{Mathematical representation} \quad
Mathematical representations signify the rigorous description of the abstract structures and patterns of human poses. The parameterized 3D human pose is original representation, which is derived from motion capture data. Mathematically, these parameters are mapped to different mathematical spaces and abstracted into different distributions whose features are easy to be extracted by the network. 
As is shown in Tab \ref{tb math}, representative papers in this category are: SRNN \cite{SRNN}, AGED \cite{AGED}, QuaterNet \cite{QN}, LTD \cite{LTD}, HMR \cite{HMR}, MGCN \cite{MGCN}, ARNet \cite{ARNET}, LDR \cite{LDR}, LPJP \cite{LPJP}, HRI \cite{HRI}, JDM \cite{Su}, SGRU \cite{SGRU}. MT-GCN \cite{MTGCN}, MPTC \cite{liu2021iccv}, MST-GNN \cite{MSGNN}, MMA \cite{HR4}.

\begin{table}[width=\linewidth,cols=3,pos=h]
\caption{Human pose representation methods table.}
\label{tb math}
\begin{tabular*}{\tblwidth}{@{} LLL @{} }
\toprule
Categories &  Methods & Years\\
\midrule
\multirow{5}{*}{Physical representation} 
    & SkelNet \cite{SKEL} & 2019\\
    & DSR \cite{DSR},  & 2019\\
    & CHA \cite{CHA} & 2020 \\
    & MGCN \cite{MGCN} & 2020\\
    & AM-GAN \cite{AMGAN} & 2021\\
\\
\multirow{17}{*}{Mathematical representation} &SRNN \cite{SRNN} & 2016\\
    & AGED \cite{AGED} & 2018\\
    & QuaterNet \cite{QN} & 2018\\
    & LTD \cite{LTD} & 2019\\
    & HMR \cite{HMR} & 2019\\
    & MGCN \cite{MGCN} & 2020\\
    & HRI \cite{HRI} & 2020\\
    & ARNet \cite{ARNET}&2020\\
    & LDR \cite{LDR} & 2021\\
    & LPJP \cite{LPJP} & 2021\\
    & TrajectoryCNN \cite{Tracnn} & 2021\\
    & JDM \cite{Su} & 2021\\
    & MMA \cite{HR4} & 2021\\
    & MPTC \cite{liu2021iccv} & 2021\\
    & SGAN \cite{SGRU} & 2021\\
    & MT-GCN \cite{MTGCN} & 2021\\
    & MPTC \cite{liu2021iccv} & 2021\\
    & MST-GNN \cite{MSGNN}& 2021\\
\hline
\end{tabular*}
\end{table}

\subsection{Network structure design}
As a major characteristic of deep learning approaches, automatic extracting and learning features from data is of considerable importance. Further, the quality of features is close to the network structure. Likewise, Network structure design approaches aim to extract the efficient features to enhance human motion prediction.
There are mainly three kinds of methods in terms of major network structures: \textit{RNN-based methods}, \textit{CN-based methods}, \textit{GAN-based methods}. 

RNN-based methods are a class of structures that are mainly composed of recurrent neural networks (RNNs) \cite{oRNN}. Currently, LSTM \cite{LSTM}, GRUs \cite{GRU} and their variants are methods that are widely used in human motion prediction due to its preponderance in processing sequential issues.
CN-based methods rely on multiple convolutional networks, such as graph convolutional networks (GCNs) \cite{LDR, LTD, MGCN}, temporal convolutional networks (TCNs) \cite{LDR}, and the variants of traditional convolutional neural networks (CNNs) \cite{Cs2s, CHA}. CN-based methods do well in capturing spatial dependencies. Moreover, they also perform well in capturing temporal dependencies by the rising of TCNs.
GAN-based methods profit from the generative adversarial networks (GANs) \cite{oGan}. These are always applied to synthetic data generation and probabilistic prediction by DNNs and a kind of novel algorithm that is utilized in network learning.

\textbf{RNN-based Methods} \quad
With the examples of the successful applications in voice recognition \cite{VR, VR1}, machine translation \cite{machine,mtr1}, and sequential prediction \cite{RNNP, SP1}, RNNs (LSTM, GRU, and their variants) have become a widely used framework for the human motion prediction task.
It is normally deemed as sequence-to-sequence (seq2seq) prediction tasks, where RNNs are adopted in solving. However, for the human motion prediction, a major distinction from other seq2seq tasks is that the human body is a human kinematics system with high constraints. 
Therefore, some RNN structures particularly for human motion prediction are proposed.  
As shown in Tab \ref{tb network}, these kinds of methods include: LSTM-3LR \cite{Res-gru}, ERD \cite{ERD},  Res-GRU \cite{Res-gru}, SRNN \cite{SRNN}, DAE-LSTM \cite{DAE}, AGED \cite{AGED}, MHU \cite{MHU}, QuaterNet \cite{QN}, SkelNet \cite{SKEL}, DSR \cite{DSR}, HMR \cite{HMR}, RNN-SPL \cite{RNN-SPL}, VGRU \cite{VGRU}, C-RNN \cite{C-RNN}, PVRED \cite{PVRED}, BNN \cite{BNN}.

\textbf{CN-based methods} \quad
In a general way, RNNs seem to be a natural choice for sequential prediction, but convolutional approaches are also employed in this task \cite{TCGAN, TCN, TCN1}. Convolution networks have their inherent advantage in capturing spatial dependencies. Thereby, convolutional seq2seq methods are well adopted. Besides, Graph Convolutional Networks (GCNs) are adaptable for presenting the human skeleton as a graph, which makes GCN widely utilized in human pose representation. Additionally, widespread recognition of the Temporal Convolution Networks (TCNs), made the CN-based methods more widely accepted for solving the sequential problem. These methods include: TE \cite{TE}, PAML \cite{PAML}, CHA \cite{CHA}, LDR \cite{LDR}, MGCN \cite{MGCN}, LTD \cite{LTD}, LDR \cite{LDR}, MoPredNet \cite{MoP}, TrajectoryCNN \cite{Tracnn}, NAT \cite{NAT}, C-seq2seq \cite{Cs2s}, Q-DCRN \cite{QDCRN}, MT-GCN \cite{MTGCN}, LMC \cite{MGCN}, MST-GNN \cite{MSGNN}, DA-GNN \cite{DA-GAN}, JDM \cite{Su}.

\textbf{GAN-based Methods} \quad
GANs have shown impressive performance in various tasks \cite{GAN1, GAN2, GAN3}. It not only provides a kind of novel learning algorithm but also offers a vital alternative to generative models. Empirically, two main strategies are deployed by leveraging GANs: promoting GANs framework performance and researching how to improve human motion prediction by these frameworks. These methods include: HP-GAN \cite{HP-gan}, BiHMP-GAN \cite{Bigan},STMI-GAN \cite{STMI-GAN},GAN-poser \cite{Gan-poser},AM-GAN \cite{AMGAN},TC-GAN \cite{TCGAN}, AGED \cite{AGED}, ARNet \cite{ARNET}, SGAN \cite{SGRU}.

\begin{table}[width=1\linewidth,cols=3,pos=h]
\renewcommand{\arraystretch}{0.4}
\caption{Convolution Network methods table}
\label{tb network}
\begin{tabular*}{\tblwidth}{@{} LLL @{} }
\toprule
Categories &  Methods & Years\\
\midrule
\multirow{15}{*}{RNN-based methods} & LSTM-3LR \cite{Res-gru}& 2015\\
    & SRNN \cite{SRNN} & 2016\\
    & Res-GRU \cite{Res-gru} & 2017\\
    & DAE-LSTM \cite{DAE} & 2017\\
    & AGED \cite{AGED} & 2018\\
    & MHU \cite{MHU} & 2018\\
    & QuaterNet \cite{QN} & 2018\\
    & SkelNet \cite{SKEL} & 2019\\
    & DSR \cite{DSR} & 2019\\
    & HMR \cite{HMR} & 2019\\
    & RNN-SPL \cite{RNN-SPL} & 2019\\
    & VGRU \cite{VGRU}, & 2019\\
    & C-RNN \cite{C-RNN} & 2020\\
    & PVRED \cite{PVRED} & 2021\\
    & BNN \cite{BNN} & 2021\\
\\
\multirow{16}{*}{CN-based methods} & TE \cite{TE} & 2017\\
    & C-seq2seq \cite{Cs2s} & 2018\\
    & PAML \cite{PAML} & 2018\\
    & CHA \cite{CHA} & 2019\\
    & LTD \cite{LTD} & 2019\\
    & LDR \cite{LDR} & 2020\\
    & MGCN \cite{MGCN} & 2020\\
    & MoPredNet \cite{MoP} & 2020\\
    & TrajectoryCNN \cite{Tracnn} & 2021\\
    & NAT \cite{NAT} & 2021\\
    & Q-DCRN \cite{QDCRN} & 2021\\
    & MT-GCN \cite{MTGCN} & 2021\\
    & LMC \cite{MGCN}, & 2021\\
    & MST-GNN \cite{MSGNN} & 2021\\
    & DA-GNN \cite{DA-GAN} & 2021\\
    & JDM \cite{Su} & 2021\\
\\
\multirow{9}{*}{GAN-based methods} & HP-GAN \cite{HP-gan} & 2018\\
    & AGED \cite{AGED} & 2018\\
    & BiHMP-GAN \cite{Bigan} & 2019\\
    & STMI-GAN \cite{STMI-GAN} & 2019\\
    & GAN-poser \cite{Gan-poser} & 2020\\
    & ARNet \cite{ARNET} & 2020\\
    & AM-GAN \cite{AMGAN} & 2021\\
    & TC-GAN \cite{TCGAN} & 2021\\
    & SGAN \cite{SGRU} & 2021\\
\hline
\end{tabular*}
\end{table}

\subsection{Prediction target}

Human motion prediction mainly contains two targets. The first purpose is to generate motion predictions that are close to the Groundtruth. Simultaneously, the future is not deterministic in the real world, which leads to the second purpose that is to generate a variety of possible motion futures.
In a word, these approaches are segmented into two classifications: \textit{probabilistic prediction} and \textit{deterministic prediction}.

\textbf{Probabilistic prediction} \quad
The future is not deterministic. So, the same prior poses could lead to multiple possible future motions.
To generate probabilistic motion prediction, the random noise vector is aggregated to the prior pose sequence or latent variables in the prediction process. 
Besides, a meaningful network learning mode is gained to upgrade the noise.  
Conclusively, as shown in Tab \ref{tb pro}, these methods include: HP-GAN \cite{HP-gan}, Dlow \cite{Dlow}, GAN-poser \cite{Gan-poser}, Mix-and-Match \cite{mix},BiHMP-GAN \cite{Bigan}, PVRED \cite{PVRED}, SGRU \cite{SGRU}.

\textbf{Deterministic prediction} \quad
Most human motion prediction methods are based on deterministic models. It is usually regarded as a regression task that produces only one outcome from a prior pose sequences.
A large proportion of the methods mentioned earlier are using deterministic prediction models, which are also used by some newly emerged methods with only a few applications in this field. We summarize these methods in this part.
\begin{table}[width=1\linewidth,cols=3,pos=h]
\renewcommand\arraystretch{1}
\caption{Prediction target methods table.}
\label{tb pro}
\begin{tabular*}{\tblwidth}{@{} LLL @{} }
\toprule
Categories &  Methods & Years\\
\midrule
\multirow{7}{*}{Probabilistic prediction} & HP-GAN \cite{HP-gan}& 2018\\
    & BiHMP-GAN \cite{Bigan} & 2019\\
    & GAN-poser \cite{Gan-poser} & 2020\\
    & Mix-and-Match \cite{mix} & 2020\\
    & Dlow \cite{Dlow} & 2020\\
    & PVRED \cite{PVRED}& 2021\\
    & SGAN \cite{SGRU} & 2021\\
\\
Deterministic prediction & others & 2015-2021\\
\hline
\end{tabular*}
\end{table}

\section{Methods Description}
In this section the detailed analysis of different methods in different categories is carried out on the basis of the  taxonomy shown in Fig.\ref{FIG:methods}. 

\begin{figure*}
	\centering
		\includegraphics[scale=.62]{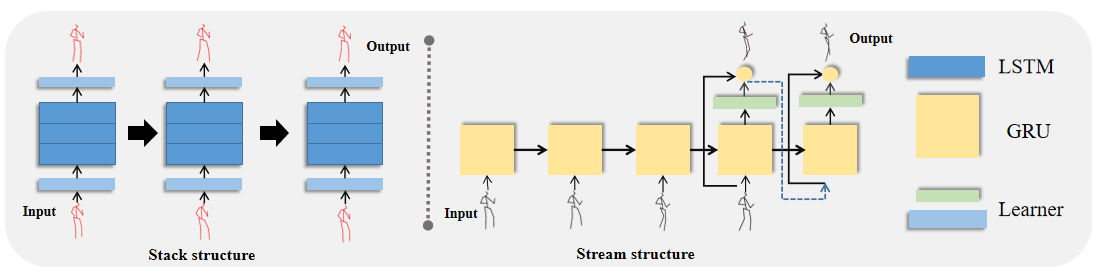}
	\caption{The basic RNNs prediction methods. These basic networks are mainly stack structure \cite{ERD} and stream structure \cite{Res-gru}.}
	\label{FIG:RNN}
\end{figure*}

\subsection{Physical representation}
The physical representation of human poses follows the principles of human motion and the basic structure of the human body. 
These representations can be understood as a class of a prior anatomical knowledge and constraints that help the network to comprehend the input information and extract features efficiently.
Generally, many existing traditional approaches \cite{Res-gru, ERD} usually utilize natural 3D joint positions to display the human skeleton. These methods mainly rely on the network performance but neglect the logical mining of human poses mocap data. 
With the progress of human motion prediction by utilizing physical representation, gradually, researchers recognize the usefulness of physical representation.
Statistically, this category can be sorted into two strategies: \textit{human body structures} and \textit{human motion laws}. 

\textbf{Human body structures} \quad 
As a representative for the first strategy, researchers attempt to optimize human body structure or adopt a new approach to learn the structure of the human body.
Guo et al. \cite{SKEL} argued to learn local structure representations in different kinematic chains in different ways. Specifically, instead of taking the holistic human pose as input, it was divided into five non-overlapping parts according to the connatural kinematic chains (the central torso, the left/right arms, and the left/right legs). This method demonstrated that paying attention to the local dynamic patterns was significant for the network to understand human motion. 
Analogously, Liu et al. proposed a new GAN framework named Aggregated Multi-GAN (AM-GAN) \cite{AMGAN}. AM-GAN modeled the motion of the central spine and four limbs. Then, the final whole human motion was aggregated over them. Nevertheless, it not only predicted future motion accurately but also can complete the tasks of controlled human motion prediction finely.
Further, Li et al. \cite{CHA} designed a convolutional network (CHA) for human motion prediction. The human pose was graded into three levels which respectively stood for joint links, five kinematic chains of human body, and the bilateral zygomorphic parts of human body. In this way, CHA can integrate multiple human body constraints into feature extraction. 
Mathematically, this strategy can be described as follows:
\begin{equation}\label{eq3.1.1}
    c_{N+T} = f_N(c_N), p_{t+T} = (c_{1+T},c_{2+T},...,c_{N+T})
\end{equation}
where $N \in N^*$, $c$ indicates kinematic chain, $p$ indicates single pose. In general, when $N$ is 5, $c$ represents the kinematic chain. If $N$ is 2, $c$ represents two parts of human body.

This type of strategy artificially attaches a prior knowledge related to the structure of the human body to the framework. However, it is relatively superficial compared with other kinematic knowledge. Moreover, networks may perceive only dimensional changes. Therefore, in terms of performance, such methods can improve the prediction to some extent.

\textbf{Human motion laws}\quad 
For the second strategy, human motion is typical of mechanical motion, that is, a part of the human body is relative to another part of the body (including the original part) in terms of space and time displacement, e.g. velocity and acceleration. 
Following this line, Wang et al. \cite{DSR} focused on that the motion of human body can be represented by velocity and acceleration instead of position. In the RNNs frameworks, Wang et al. proved that these frameworks were easier to fit the data represented by velocity and acceleration. 
However, this performance improvement is limited to short-term prediction (less than $400$ milliseconds) and does not effectively ameliorate the intrinsic defects of RNNs.
Similarly, Li et al. \cite{MGCN} considered learning richer motion dynamics for human motion prediction. Therefore, the difference operators were deployed to extract multiple physical information: positions $X$, velocities $ \Delta $ $X$, and accelerations $\Delta^2 X$. Summarily, the second strategy can be described as follow:
\begin{equation}\label{eq3.1.2}
    \Delta^nX_t = \Delta^{n-1}X_t - \Delta^{n-1}X_{t-1}, 0<n<3
\end{equation}

These methods, which follow the motion laws, are worthy of study. These data processed by motion laws, such as velocity and acceleration, comprise richer kinematic knowledge, naturally containing inter-frame information and more stable data variations. Deeper research on the usability of such methods is needed.

\textbf{Summary of physical representation} \quad 
Physical representation plays a seminal role in prediction performance, both in terms of human structure and human motion laws. However, in both categories, the human body structure is shown to be less effective than human motion laws. The reason is that the former is explored only in terms of the shape of the human pose vector, whereas the latter provides richer information on physics that is not confined to a single frame. Therefore, attempts can be made to incorporate more physical knowledge into the field of human motion prediction.

\subsection{Mathematical representation}
The human pose has its immanent structure in the scientific research, which is mostly presented as the 3D skeleton representation captured by a motion capture system. 
The mathematical representation of human pose is a general way to strictly describe the abstraction of the human pose. It allows experts in this domain to inject their prior knowledge into the learning process of these human poses. 
Existing mathematical representations can be sorted into three directions, \textit{Graph}, \textit{Motion trajectory}, and \textit{Mathematical encoding}.

\textbf{Graph}\quad 
The joints and links between adjacent joints of the skeleton representation (a human skeletal kinematic tree) naturally remind the researchers of graph \cite{LTD, HRI}.
Graph $G(V,E)$ is a data format and also an encoding mode that can be used to represent social networks \cite{graph1,graph2}, communication networks \cite{graph3, graph4}, protein molecular networks \cite{graph5,graph6}, etc. The nodes $E$ in the graph represent individuals in the network, and the linked edges $V$ represent connections between individuals. 

Inspired by this, Yan et al. \cite{SRNN} first constructed a spatial-temporal graph to define the human body. However, these graphs were manually designed, thus limiting the flexibility of the algorithm.
Considering this, in \cite{LDR}, Cui et al. designed a trainable graph embedded into the network. The adjacency matrix that represented the weights of the connected joint parts was set as the model parameters. A similar method was also used in \cite{MTGCN, Su}.
Analogously, Mao et al. \cite{LTD} also created a new method to learn an automatic-connected graph. Innovatively, it proposed the non-restrictive joint dependencies without the constraints of the kinematic tree or the convolutional kernel size. 
Nevertheless, such constructions are tantamount to treating pose sequence roughly as routine data, losing sight of significant kinematic constraints, which leads to unstable training. 
This strategy was also adopted in their method \cite{HRI}. 
To overcome this drawback, they put forth a new method \cite{HR4} that considered the kinematic constraints from three levels of the human body. The first level was the full-body, the second was the parts of human body, and the third was the individual joints.

Parallel to this work, considering the human kinematic structure, Li et al. proposed two multi-scale graphs to model the human body. 
In \cite{DMGNN}, a multi-scale graph was presented to capture multi-scale features to predict future human motions. It contained two kinds of sub-graphs, namely single-scale graphs and cross-scale graphs, which respectively connected the human body components at the same scale and cross-scale. 
Subsequently, they also proposed a multi-scale spatial-temporal graph (st-graph) \cite{MSGNN} to comprehensively model human motion. The characteristics of the st-graph are trainable, multi-scaled, and decomposable.  
To be specific, the st-graph can be decomposed respectively into a spatial graph and a temporal graph, both are trainable.
Meanwhile, Liu et al. \cite{liu2021iccv} devised a semi-constrained graph to explicitly encode skeletal connections and used prior knowledge (such as limb mirror symmetry tendency and cross sides synchronization tendency). This kind of graph proves that the combination of adaptive learning and constraints is beneficial to training.

\textbf{Motion trajectory}\quad 
The human motion trajectory is the spatial motion feature composed of the path taken by a part of the human body from the beginning position to the end. In recent years, the progress of the trajectory space for human motion prediction has aroused researchers' interest in modeling human motion in the trajectory space rather than in traditional motion space. 

Based on this, Mao et al. \cite{LTD, HRI, HR4} presented human motion in trajectory space for several times. 
Parallel to these, Su et al. \cite{Su} also adopted the same trajectory space of a joint. Mathematically, these methods can be formulated as:
\begin{equation}\label{eq3.2.1}
    T_j = (t_{j,1},t_{j,2},t_{j,3},...,t_{t,N})
\end{equation}
where $j$ denotes $j^{th}$ joint, $N$ denotes the number of frames. 
As the most commonly used trajectory representation, it is favorably received by researchers. 
Differently, Liu et al. \cite{liu2021iccv} proposed a new trajectory format, which consisted of the position displacement of the adjacent frames and motion direction. Similarly, in  \cite{Tracnn}, a novel representation was invented to build the correlations between the local and global space, which were respectively representative of the same part and different parts in the human motion sequence. 
The motion trajectory is smoother compared with traditional space, which contributing to more stable data variation. As a result, this will bring more convenience to the network training.

\textbf{Mathematical encoding}\quad
It is clear that reasonable constraints are quite conducive to the network to learn the features. 
A mathematical expression is a formulaic constraint that is in a sense easier to program.
Thereby, finding an effective mathematical way to encode the human pose is undoubtedly of great significance to human motion prediction. However, this process requires much richer prior knowledge. To inspire researchers, the existing mathematical methods are summarized and then divided into the following two categories: \textit{algebra encoding} and \textit{differential encoding}.

\quad \textit{Algebra encoding}\quad 
Algebraic encoding deals with the problem of human motion prediction by studying the number, relationship and structure of human poses. Common types of algebraic structures are groups, loops, fields, modules, etc. Generally, each pose is described as a human body joint positions union or a 3D-joint rotations union.
In most cases, human joints are represented as some algebraic form that captures dependencies more easily.
In \cite{QN}, they represented joint rotations with quaternions, which lie in $R^4$. 
The quadratic representation is not plagued by discontinuities and singularities. It is more stable with respect to the variation of number and is computationally more efficient, with enormous potential in studying human motion prediction.
Another representation is to utilize the Lie Group \cite{Lieg}, which is a Riemannian manifold structure. 
In \cite{AGED} and \cite{HMR}, they both adopted this method to encode the human pose. Differently, Gui et al. \cite{AGED} only made use of the Lie groups to characterize the relative 3D geometry for a single joint, but Liu et al. \cite{HMR} characterized it between two successive bones. Regarding effectiveness, Liu et al.'s method was more suitable for modeling with the Lie group. 
After the introduction to the generality of trajectory representation above, efficient ways of encoding trajectory have emerged.
The strength of trajectory representation lies in its smooth representation of the object's motion.
Discrete Cosine Transform (DCT) \cite{DCT} is widely used in many methods \cite{LTD, HRI, HR4, ARNET, LPJP} for encoding trajectory. By discarding the high frequencies, it can generate a more compact representation, which can transform the 3D human joints coordinates into smoother motion state, predestining its popularity.
Distinctively, the phase space trajectory representation was proposed by Su et al. \cite{Su}. They employed the joint instantaneous displacements from the joint displacement of adjacent frames to form the joint trajectory. 

\quad \textit{Differential encoding}\quad 
The differential method itself is an advanced mathematical method, which sets, higher requirements on researchers' ability. Therefore, at first, researchers use differential methods only for simple problems such as \cite{MGCN} and \cite{MSGNN}. A difference operator is utilized to compute the differences of observed poses. These differences can reflect richer dynamic information (e.g., velocities, and accelerations) for the network. They cannot play a decisive role in solving the problem. 
In \cite{SGRU}, Lyu et al. proposed a novel modeling method that utilizes stochastic differential equations \cite{sde} to model human pose. The motion profile of each skeletal joint was formulated as a basic stochastic variable and modeled with the Langevin equation \cite{lang}. This approach merely required a basic and lightweight network to achieve good results.

\textbf{Summary of mathematical representation} \quad 
The mathematical representation is a rigorous abstraction of the human pose. From a practical perspective, effective mathematical representations are quite helpful in understanding the principles, for both humans and machines. This is definitely a pretty indispensable way to meliorate predictive performance.

\subsection{RNN-based methods}
Human motion prediction problems are often treated as seq2seq prediction tasks. RNNs are widely recognized for their excellent performance on such tasks, which has inspired many researchers to utilize RNN-based methods to investigate human motion prediction tasks. As is shown in Fig.\ref{FIG:RNN}, two basic RNNs (stack structure and stream structure) are deployed on it. 

Primitively, two architectures LSTM-3LR \cite{Res-gru} and ERD \cite{ERD} were proposed. Their main frameworks were concatenated LSTM units. The difference that made ERD perform better was the non-linear space encoder. However, these kinds of networks were still trapped in accumulated errors, discontinuity in initial frames, which quickly produce unrealistic human motion.
Therefore, the res-GRU \cite{Res-gru} was designed to solve these issues. They modeled the velocity of joints to represent the human body. For the network, they employed the linear layer to encode the pose features and decode the hidden states. It firstly demonstrates the efficiency of the velocity modeling. The noise was combined with the input sample during training, which prompted a more robust network to alleviate drifting. However, this strategy brought about difficult training sessions.
In \cite{SRNN}, Jain et al. proposed structural RNNs (SRNNs), in which the manually designed spatial-temporal graph was later employed. Noise was also employed to SRNNs.
Then, the DAE-LSTM \cite{DAE} combined a LSTM-3LR with a dropout auto-encoder to model temporal and spatial structures. 
The above methods have laid a solid foundation for subsequent researches \cite{QN, SKEL, DSR, RNN-SPL, BNN, AGED, PVRED,C-RNN}. It is worth mentioning that, some methods have made innovative design in network structure. 

The improved performance inspired us to focus on the kinematics during the network design. Some researchers find that not all joints are involved in human actions. In other words, some joints are motionless during the motion. 
Based on this, in \cite{MHU}, authors proposed a modified network (HMU) for human motion prediction without motionless joints for long-term prediction. Specifically, they designed a novel gate structure to filter the motionless joints. What’s more, an attention module was utilized to concentrate on these kinds of motions.
Then, Liu et al. \cite{HMR} proposed a hierarchical recurrent network (HMR). To avoid the impact of recent frames, the input of the network was set as a single pose. To capture more long-term dependencies, in the recurrent units, adjacent joints and frames can be encoded concurrently.
Further, a new RNN framework was also presented by them, which was named AHMR \cite{liu2022}. It not only can concurrently model the local and global context but also an attention module was employed to assist in updating the global context. Moreover, two effective loss functions were designed in this work. In this way, a longer-term and more real human motion sequence can be generated.
Respectively, a hierarchically decomposed network was presented in \cite{VGRU}. To be more specific, they proposed a coarse-to-refine network to generate the states of each stage. In the coarse stage, the guide vectors were created as coarse states for the next stage. Then, the actual pose outputs were gained from the closed-loop predictions.

\textbf{Summary of RNN-based methods} \quad 
The RNN-based methods, as always, show marvelous handling of timing problems. However, human motion prediction is not purely a temporal problem, as it has kinematic and human anatomical constraints. Therefore, the development of such methods has also moved from exploiting the performance of recurrent networks to incorporating effective constraints. Even so, the inherent problems of RNNs continue to plague researchers to varying degrees, which needs to be addressed on an ongoing basis.

\subsection{CN-based methods}
Human motion involves both spatial and temporal correlations. CN-based methods have a natural advantage in capturing spatial dependencies. However, traditional CNNs are stuck in limitations in processing standard 2D grid representations. Thanks to the GCNs, the spatial correlations can be captured efficiently. With the introduction to TCNs, the ability of CN-based methods to deal with temporal correlations becomes prominent. In summary, the CN-based methods include three sorts: \textit{conventional convolution methods}, \textit{GCNs methods}, and \textit{TCN methods}.

\textbf{Conventional convolution methods}\quad
Traditional CNNs are widely used in human motion prediction for their inbuilt ability to capture spatial dependencies.
In \cite{TE}, a convolutional layer was designed to encode different time scales. It was only utilized to capture the local time scales. 
In QuaterNet \cite{QN}, dilated convolutions was utilized in the network to capture long-term temporal dependence with the hierarchical input poses.
However, these methods can only capture single and temporal information.
Further, a hierarchical structure of CNN \cite{Cs2s} was employed in the network to capture spatial and temporal dependencies. The convolutional structure was utilized in the encoder to obtain a long-term hidden state, which was fed to the decoder to generate the human poses.
Then, Li et al. \cite{CHA} designed a convolutional hierarchical autoencoder framework. Concretely, hierarchical topology was employed to represent the human body tree structure, and 1D convolutional layers were embedded to leverage these constraints to encode each node. 
A novel framework named TrajectoryCNN \cite{Tracnn} was designed to predict future poses. It introduced a new type of trajectory space for spatio-temporal dependencies, which included a variety of local-global and spatio-temporal features, that can be captured easily by CNNs.

\begin{figure*}
	\centering
		\includegraphics[scale=.58]{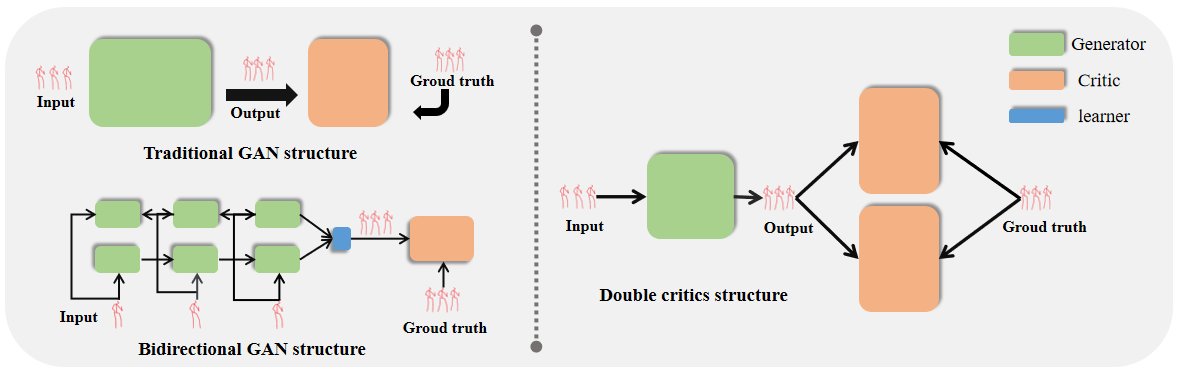}
	\caption{The basic GANs prediction structure. Structurally, there are two kinds of frameworks, traditional GAN structure \cite{HP-gan}, bidirectional GAN structure \cite{Bigan}, and double critics structure \cite{AGED}.}
	\label{FIG:GAN}
\end{figure*}

\textbf{GCNs methods}\quad
GCN actually does the same thing as CNN, which is a featured extractor, except its object is graph data. It is an ingeniously designed method for extracting features from graph data. With this method, the features extracted can be used to achieve a wide range of purposes, such as node classification, graph classification, link prediction, and incidentally, graph embedding and so on. In Fig.\ref{FIG:GAN}, two classic GANs structures are shown.

In \cite{LTD}, a human pose was encoded as a graph structure, which followed the principle of connecting every neighboring joints. Moreover, they proposed a new GCN to automatically connect the graph instead of manual defining. 
Then, a novel graph network was proposed as a generator in the GANs \cite{LDR}. A dynamic learning graph was also used but it was different from a generatic one. Because it not only can explicitly learn the natural joint pairs but also implicitly connect geometrically separated joints.   
In \cite{DMGNN}, Li et al. designed a novel GCN named DMGNN which contained a dynamic multi-scale graph to represent the human skeleton structure. The multi-scale graph can comprehensively model the internal relations of a human body. It also can be used during dynamic across network layers learning. To generate future poses, a proposed graph-based gate recurrent unit was employed for this task.
In order to generate high-fidelity human motion predictions from incomplete observations, \cite{MTGCN} also proposed a novel multi-task graph convolutional network (MT-GCN), which included a shared context encoder (SCE). 
In addition to graph structure, a temporal self-attention mechanism was selected to the most related information from the whole sequence to repair the corrupted pose. This kind of multi-scaled method was also present in \cite{MGCN} to capture the correlation of body components.

\textbf{TCN methods}\quad
In recent years, TCNs display their outstanding performance in dealing with sequential tasks. TCNs are mainly composed of Causal Convolution and Dilated Convolution. Causal convolution makes judgments based on historical and current information, rather than future information, which is in line with the characteristics of time sequence. Atrous convolution is not restricted by the size of the convolution kernel, which enables CNN to process longer time series with certain parameters. Owing to these, TCNs are also extensively utilized in human motion prediction.
In \cite{MoP}, a novel approach named Motion Prediction Network (MoPredNet) for few-short human motion prediction was proposed. MoPredNet can be adapted for predicting new motion dynamics using limited data, and it delicately captured long-term dependency in motion dynamics. Thereinto, a deformable Spatio-Temporal Convolution Network (DSTCN) was proposed to adaptively model sub-motion dynamics and spatial correlation in the entire motion sequence to capture long-term dependency for motion generation. 
Further, Cui et al. presented the residual TCN (ResTCN) in the TCGAN \cite{TCGAN}. It was utilized as the generator for motion prediction. ResTCN was employed to model sequential prediction tasks due to its few parameters and high efficiency.

\textbf{Summary of CN-based methods} \quad 
In summary, all three methods effectively employ convolutional modules to handle the task of human motion prediction. Traditional methods focus on the ability to use convolutional networks for spatially dependent capture.
GCNs benefit from the effective a priori knowledge-graph that can efficiently represent the relationships between each human body part.
TCNs have become popularized when dealing with its outstanding performance for sequential problems, even surpassing RNNs.

\subsection{GAN-based methods}
On account of the extensive use of generative adversarial networks (GANs) \cite{oGan}, it provides many new directions for human motion generation by the probability density function and network learning algorithms. In human motion prediction task, GANs also contribute to substantial progress. 

As the first application of GANs in human motion prediction, Barsoum et al. \cite{HP-gan} proposed a novel sequence-to-sequence model for probabilistic human motion prediction, trained with a modified version of improved Wasserstein generative adversarial networks (WGAN-GP). 
Different from previous works, HPGAN \cite{HP-gan} represented the input of the network as probability distribution during the training. The probability density function can assist to predict multiple future pose sequences by inputting the same prior sequence with a different vector $z$. This is also a common practice to produce diverse futures. However, they failed to assess expressiveness of such a generative approach against deterministic counterparts.
After this work, a new probabilistic generative approach called Bidirectional Human motion prediction GAN, or BiHMP-GAN \cite{Bigan} was designed. 
Similarly, BiHMP-GAN was also able to generate probabilistic future motions with the same input incorporating a random vector $r$ from predefined distribution. Likewise, a bidirectional GAN framework was designed to avoid the mode-collapse. In this method, the discriminator was also trained to regress the vector $r$. Further, BiHMP-GAN made the comparison with other deterministic approaches. 
Also as a bidirectional GAN framework, GAN-poser \cite{Gan-poser} was employed to avert mode collapse and to further regularize the training. 
Similar to the previous method, it also utilized a random factor to generate multiple future motions. In spite of being in a probabilistic framework, the modified discriminator architecture allowed considering the randomness of the prediction. In addition, a new evaluation was provided in this work for assessing the prediction results.
Obviously, these bidirectional GAN frameworks performed better than general GANs.

Parallel to these works, inspired by the adversarial training mechanism, AGED \cite{AGED} presented a novel GAN framework with two global recurrent discriminators. One discriminator was utilized to promote the fidelity of the generation sequence, the other discriminator was trained on the joint-level to guarantee the continuity of generated future sequences.
STMI-GAN \cite{STMI-GAN} also adopted the adversarial learning for spatio-temporal tensor of 3D skeleton coordinates over long periods of time. 
Dexterously, Adversarial Refinement Network (ARNet) \cite{ARNET} designed a novel adversarial error augmentation. Different from the normal adversarial learning, the discriminator was utilized as a middle module to produce the prediction errors, which were transferred to the refinement module.
Interestingly, unlike previous works, Lyu et al. \cite{SGRU} utilized the GANs to simulate path integrals for solving the stochastic differential equations and predict future motion profiles. 

However, it should not be overlooked that it was challengeable to train GANs. Since it was difficult to reach the Nash equilibrium between the generator and discriminator, Cui et al. \cite{TCGAN} presented a new GAN with spectral normalization to avoid mode collapse. 
There's another strategy called AMGAN \cite{AMGAN}, which was designed from a composite GAN structure, consisting of local GANs for different low-dimensional body parts and a global GAN for the high-dimensional whole body. This method proved that dimension reduction can efficaciously improve the training efficiency of GANs.

In summary, the strategy of utilizing GANs mainly can be divided into two categories. ($i$) Being used as a learning algorithm to help with network learning. ($ii$) Utilizing data distribution to generate multiple prediction targets. GANs, as a network with noticeable advantages and disadvantages, can also give rise to certain challenges for researchers' work.

\begin{figure*}
	\centering
		\includegraphics[scale=.58]{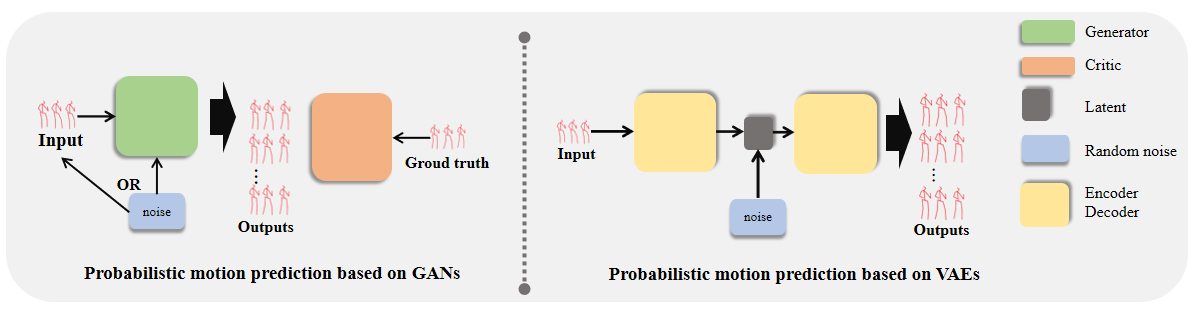}
	\caption{The basic probabilistic motion prediction structure. Structurally, there are two kinds of frameworks, GANs structure \cite{HP-gan, Bigan}, and VAEs structure \cite{Dlow}.}.
	\label{FIG:prob}
\end{figure*}

\subsection{Probabilistic motion prediction}
The inherent stochastic nature of motions always besets the machine to understand and predict human motion. What’s more, for human future motions, the longer the time period is, the more uncertain the result will be. Even the subject of motion cannot be sure about his/her movement over a long period of time. Therefore, probabilistic motion prediction is  indispensable for the research of human motion prediction.
According to the implementation means and network framework of existing methods, these are mainly divided into two categories: \textit{GAN-based methods} and \textit{VAE-based methods}. As shown in Fig.\ref{FIG:prob}, it shows the basic probabilistic motion prediction structures.

\textbf{GAN-based methods}\quad 
As a common strategy for the GAN approaches, they generate probabilistic future motions by merging the random noise into the observed pose sequence.
The HPGAN \cite{HP-gan} initially proposed to utilize the GAN framework to fulfill the probabilistic human motion prediction task. It generated multiple future pose sequences only from a sing observed pose. In this strategy, a random vector $r$ was combined with the prior pose sequence. Each random vector can assist to produce a possible future motion. This idea was also expressed in \cite{SGRU, Gan-poser}. Nevertheless, the issue of discontinuity of motion cannot be overcame by this method.
To hit the target, BiHMP-GAN \cite{Bigan} utilized a single random vector to generate the entire sequence. This vector was employed to alter the initialization of the decoder and concatenated with a pose embedding at each iteration of the RNNs. By relying on concatenation as a method to fuse the condition and the random vector, these two methods contained parameters that were specific to the random vector, and thus brought the model flexibility to ignore this information. 

\textbf{VAE-based methods}\quad 
The second category of probabilistic motion prediction is achieved by an additional CVAE. It is similar to GAN in that it changes the output by fusing the noise. Differently, the random noise will turn into the intermediate variable before the passing of decoder.
In \cite{mix}, directly, the human pose was utilized as a conditioning variable to combine the random noise. However, it generated low-grade human motion due to the incoherence of random noise at each step.
However, in \cite{cvae1}, the RNN decoder hidden state taking place of the pose was employed as a conditioning variate. This method effectively promoted performance but neglects the flexibility of the random vector.
Yuan et al. skillfully designed a new sampling approach named DLow \cite{Dlow}. Different from the normal random sampling, a set of learnable mapping functions was employed to produce a set of correlating latent variates only from an input of single random noise. 

In summary, probabilistic human motion prediction relies on a combination of random noise and conditional variables. Definitely, the network structure that can work on this is equally important. The existing underlying networks are GANs and VAEs. The combined progress of these two paths will unquestionably be a major boost.

\subsection{Deterministic motion prediction}
Deterministic motion prediction is the most widely applied human motion prediction means. Generally, human motion prediction is always regarded as a regression task. RNNs are acknowledged to be skilled in handling such tasks. Therefore, many researchers attempted to utilize RNNs to cope with human motion prediction. Typically, these approaches \cite{ERD, Res-gru, SRNN, DAE, QN, SKEL, DSR, BNN, AGED, PVRED, C-RNN,MHU, HMR, VGRU} effectually predict future human motions in the recurrent framework. Similarly, the feed-forward models \cite{TCGAN} are also deployed for this task. Recognizing the positive impact of kinematics on prediction, many convolutional methods \cite{CHA,Cs2s, TE,Tracnn}, especially the graph neural networks \cite{LDR, LTD, DMGNN, MSGNN, STMI-GAN, AGED,ARNET}, are utilized. 

With the popularity of attention networks, there are many proposed solid methods.
In \cite{MoP}, a novel approach named Motion Prediction Network (MoPredNet) for few-short human motion prediction was presented. Specifically, MoPredNet dynamically selected the most informative poses in the streaming motion data as masked poses. In addition, MoPredNet enhanced its encoding capability of motion dynamics by adaptively learning spatio-temporal structure from the observed poses and masked poses.
In \cite{LPJP}, a transformer-based method was proposed to solve the human motion prediction issue. This framework can concurrently capture temporal and spatial dependencies of human motion. With the employing of the global attention mechanism, the network can acquire more effective long-term dependencies.

\section{Datasets and Performance Evaluation}
It goes without saying that the design of experiments is a critical part of academic research. In this section, the datasets and the evaluation criteria of human motion prediction are expatiated.

\subsection{Datasets}
Datasets are a pivotal part of algorithm research in different fields. Generally, they play a principal role that facilitates the learning of networks and measures the performance as a common ground. Apart from this, the improved quality of datasets has also driven the field to be more valuable and challenging consequences. In particular, deep learning has drawn much more attention recently and it is helpful in part of the huge amount of data. Hence, more and more datasets are established to tackle the problems.
For human motion prediction, seven standard datasets are applied to assist learning, as shown in Tab \ref{dataset}. They are Human 3.6M(H36M) \cite{H36M}, CMU motion capture (CMU Mocap), 3D Poses in the Wild (3DPW) \cite{3Dpose}, The Archive of Motion Capture as Surface Shapes (AMASS) \cite{AWASS}, G3D \cite{G3D}, NTU RGB+D \cite{NTU}, Filtered NTU RGB+D (FNTU) \cite{FNTU}, Whole-Body Human Motion (WBHM) \cite{WB}. The above-mentioned datasets will be explicated respectively in the following part.

\begin{table*}
[width= 2 \linewidth,cols=5,pos=h]
\renewcommand\arraystretch{1.1} 
\caption{Datasets table. We collected all the datasets used in relevant papers and explained some of their attributes.}
\label{dataset}
\begin{tabular*}{\tblwidth}{@{} LLLLLLL @{} }
\toprule
Datasets & Year & Location & Number of Joints & FPS &  Sensors &\\
\midrule
H36M \cite{H36M} & 2014 & indoor & 32 & 25 & 10 Vicon T40 &\\
CMU \cite{cmu} & 2003 & indoor & 38 &  25 & 12 infrared cameras &\\
3DPW \cite{3Dpose} & 2018 & outdoor & 17 &  30 & A hand-held smartphone camera &\\
G3D \cite{G3D} & 2012 & indoor & 20 & 30 & The Microsoft Kinect device &\\
AMASS \cite{AWASS} & 2019 & indoor & 52 &  25 & OptiTrack mocap system &\\
NTURGB-D \cite{NTU} & 2016 & indoor, outdoor & 25 &  25 & Kinect V2 &\\
FNTU \cite{FNTU} & 2019 & indoor & 25 &  25 & Kinect V2 &\\
WBHM \cite{WB} & 2015 & indoor & 56 &  25 & Vicon MX &\\
\bottomrule
\end{tabular*}
\end{table*}

\textbf{Human 3.6M(H36M)}\quad  
H36M public dataset records human motion data, which includes 3D human poses and their corresponding images with 5 females and 6 males. It totally contains 3.6 Million data recorded by a vicon motion capture system with 4 different viewpoints. These poses involve 15 different scenarios of complex actions, which include directions, discussion, eating, greeting, phoning, posing, purchases, sitting, smoking, taking photo, waiting, walking, walking dog, and walking together. In addition, each scenario includes many types of asymmetries, e.g. walking with a hand in a pocket, walking with a bag on the shoulder. The pose parametrizations comprise joint positions and joint angle skeleton representations and a full skeleton consists of 32 skeletal joints. Experimentally, researchers always divided these poses into 7 subjects (S1, S5, S6, S7, S8, S9, S11) and removed duplicate points of the human pose and 25 points are retained. A down-sampling is applied to set 25 frames per second (FPS).
Datasets have been publicly available at \url{https://vision.imar.ro/human3.6m}.

\textbf{CMU motion capture (CMU)} \quad 
In 2003, Carnegie Mellon University released a public dataset CMU recorded by 12 infrared cameras. There are 41 markers taped on the human body. 144 different subjects are embodied into this dataset, such as basketball, basketball-signal, directing-traffic, jumping, running, soccer, walking, wash-window, and so on. 38 joints make up the parameterized human pose. In general, these samples are split into training set and test set in the experiment. The sequences are down sampled to meet the frame rate of 25fps.
This dataset has been publicly available at \url{http://mocap.cs.cmu.edu/}.

\textbf{3D Poses in the Wild (3DPW)} \quad  
3DPW dataset is mainly presented for wild scenes. It is a kind of large-scale publicly available dataset, which contains 60 video sequences, more than 51, 000 indoor or outdoor poses. This dataset is recorded by a hand-held smartphone camera or IMU. Generally, two actors equipped the IMU to act different actions, such as shopping, doing sports, hugging, discussing, capturing selfies, riding bus, playing guitar, relaxing. There were totally 7 actors with 18 clothing styles. Each pose consists of 17 joints. The frame rate is 30fps.
Datasets have been publicly available at \url{http://virtualhumans.mpi-inf.mpg.de/3DPW}.

\textbf{G3D} \quad
It is a gaming dataset collected with Microsoft Kinect device and Windows SDK. It provides real-time action recognition in gaming containing synchronized video, depth, and skeleton data. In total, there are 210 samples and 10 subjects performing 20 gaming actions. These actions include punch right, punch left, kick right, kick left, defend, golf swing, tennis swing forehand, tennis swing backhand, tennis serve, throw bowling ball, aim and fire gun, walk, run, jump, climb, crouch, steer a car, wave, flap, and clap. Each pose is made up of 20 defined joints. The frame rate is 30fps.
This dataset has been publicly available at \url{http://dipersec.king.ac.uk/G3D/}.

\textbf{The Archive of Motion Capture as Surface Shapes (AMASS)} \quad
AMASS dataset is a large and public dataset for parameterized human motions. It contains 15 different optical markers to record the human body. It entirely includes 344 subjects and 11265 motions within 42 hours. Habitually, 52 joints are utilized to represent a human pose.
Datasets have been publicly available at \url{https://amass.is.tue.mpg.de/}.

\textbf{NTU RGB-D} \quad
As a large-scale action recognition dataset, the NTU RGB-D records a lot of the corresponding RGB videos, depth map sequences, 3D skeletal data, and infrared videos by three Kinects from different viewpoints. 60 action classes, 56,880 actions, 40 different subjects, and 4 Million frames are recorded by the Microsoft Kinect API. Each action is made up of 25 major joints. 
This dataset has been publicly available at \url{http://rose1.ntu.edu.sg/datasets/actionre\\cognition.asp}.

\textbf{Filtered NTU RGB+D (FNTU)}\quad
FNTU is closely connected with NTU. It is specialized for human motion task due to the noisy of the human skeletal data. These noisy skeletal sequences are not suitable for pose prediction. FNTU filters the mutual actions and selects the relative forward skeleton of the human body. This dataset is composed of 18,102 samples, among which 12,001 samples are selected for training and the rest for testing.
This dataset has been publicly available at \url{https://drive.google.com/drive/folders/ \ 1bqNyIk2O \ 0NIf5Hv\%202sMfwsuPjwbpZK-n5}.

\textbf{Whole-Body Human Motion (WBHM)}\quad
WBHM is a large-scale publicly available whole-body dataset containing 3D raw data of multiple individuals and objects. It is constituted by captured raw motion data as well as the corresponding post-processed motion. This database serves as a key element for a broad variety of research questions related e.g. to human motion analysis, imitation learning, action recognition and motion generation in robotics. The motion database is comprised of motion data of a total run length of 7.68 hours. 43 different subjects (31 males and 12 females) and 41 different objects have been  included into the database. 289 motion capture experiments, defined as a combination of subject and motion type, have been conducted in our motion capture lab. The human pose consists of 56 joints.
Datasets have been available at \url{https://motion-database.humanoids.kit.edu/anthropometric table/}.

\begin{figure}
	\centering
		\includegraphics[scale=.5]{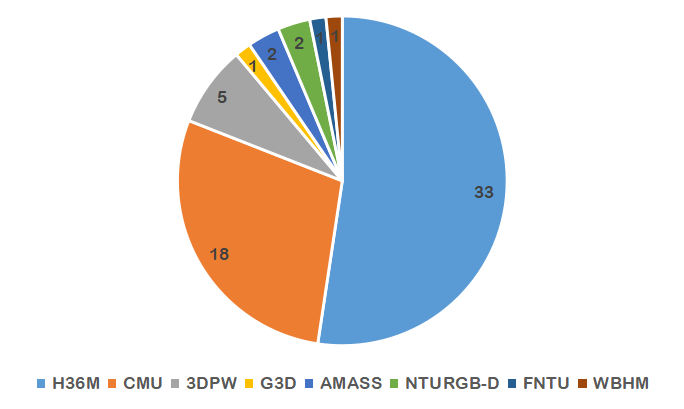}
	\caption{The datasets are used in this domain. }
	\label{diagram}
\end{figure}

As shown in Fig.\ref{diagram}, H36M dataset is the most widely used in human motion prediction. It contains a large amount of data and a wide variety of actions. Since it was used by researchers early on, H36M has been used extensively for comparison. However, other datasets are still rarely used by researchers. This shows the broad development space in this field.  

\subsection{Performance Evaluation}
It is difficult to predict future human motion, because of the high dimension and stochastic nature of human motion. In this case, a momentous challenge for human motion prediction is how to evaluate the prediction performance, which can measure the similarity between the predicted and the actual motion, and further compare with other methods. Different classes of tasks require different evaluations. To sum up, there are mainly two metrics in our tasks: \textit{Geometric accuracy metrics} and \textit{Probabilistic accuracy metrics}.  

\subsubsection{Geometric accuracy metrics}
In most circumstances, Geometric accuracy metrics are used in academic researches or application domains. 
To conclude, accurate metrics can be classified into four kinds of methods: Mean Angle Error (MAE), Mean Per Joint Position Error (MPJPE), Average Displacement Error (ADE), Final Displacement Error (FDE).

\textbf{Mean Angle Error (MAE)} \quad
MAE is an immensely popular metric for human motion prediction.  
When given the Euler angles $\hat{a}_{n,k}$, MAE is utilized as metrics. In terms of form, the MAE can be described as follow:
\begin{equation}\label{eq1}
    MAE = ({N_a}K)^{-1}\sum_{n=1}^{N_a}\sum_{k=1}^K|\hat{a}_{k,n} - a_{k,n}|
\end{equation}
where $\hat{a}_{k,n}$ denotes the predicted $k^{th}$ angle in frame $n$ and $x_{k,n}$ is the ground truth. 

\textbf{Mean Per Joint Position Error (MPJPE)} \quad
3D coordinates joints are the original representation for human body. 
MPJPE is also a widely accepted metric. 
Formally, the MPJPE can be described as follow:
\begin{equation}\label{eq2}
    MPJPE = \frac{1}{{N_j}K}\sum_{n=1}^{N_j}\sum_{k=1}^K||\hat{j}_{k,n} - j_{k,n}||^2
\end{equation}
where $\hat{j}_{k,n} \in R^3$ is the predicted $j^{th}$ 3D joint position in frame $n$ and $j_{k,n}$ is the ground truth.

\textbf{Normalized Power Spectrum Similarity (NPSS)} \quad

NPSS is proposed to evaluate the long-term predictive ability of motion synthesis models, complementing the popular mean-squared error (MSE) measurement of Euler joint angles over time.
NPSS is meant to complement MSE by addressing some of its drawbacks for application as a quantitative evaluation metric for long-term synthesis. Formally, the NPSS can be described as follow:
\begin{equation}\label{eq3}
    NPSS = \frac{\sum_i \sum_j p_{i,j}*end_{i,j}}{\sum_i \sum_j p_{i,j}}min_{x\in X} ||\hat{x}-x||
\end{equation}
where $p_{i,j}$ is the total power of $i^{th}$ feature in $j^{th}$ sequence.

\textbf{Average Displacement Error (ADE)} \quad
ADE is utilized to average $L_2$ distance over all time steps between the ground truth motion $\hat{x}$ and the closest sample. The formula is described as follow:
\begin{equation}\label{eq4}
   ADE = \frac{1}{T}min_{x\in X} ||\hat{x}-x||
\end{equation}

\textbf{Final Displacement Error (FDE)} \quad
FDE utilizes $L_2$ distance between the final ground truth pose $x^T$ and the closest sample’s final pose, which is computed as min $x\in X$. The formula is described as follow: 
\begin{equation}\label{eq5}
    FDE = ||\hat{x}^T-x^T||
\end{equation}

\subsubsection{Probabilistic accuracy metrics}

For probabilistic accuracy metrics, \cite{HP-gan} utilizes a discriminator network, whose sole purpose is to learn the probability that whether a given sequence is a valid human motion. In \cite{Dlow}, Average Pairwise Distance (APD) is used to measure sample diversity. APD average L2 distance between all pairs of motion samples to measure diversity within samples, which is computed as:
\begin{equation}\label{eq6}
   APD = \frac{1}{(K-1)K}\sum_{i=1}^{K}\sum_{j\neq k}^K||x_i - x_j||
\end{equation}
where $K$ is the number of joint and $x$ is the motion sample.

\begin{table*}
[width=2\linewidth,cols=17,pos=h]
\scriptsize
\centering
\caption{Performance evaluation (in MAE) on H36M dataset for both short-term and long-term human motion prediction.}\label{h36mmae1}
\begin{tabular*}{\tblwidth}{@{} LLLLLLLLLLLLLLLLL @{} }
\toprule
Types & 
\multicolumn{2}{L}{Directions}& 
\multicolumn{2}{L}{Discussion} & 
\multicolumn{2}{L}{Eating} & 
\multicolumn{2}{L}{Greeting}& 
\multicolumn{2}{L}{Phoning}& 
\multicolumn{2}{L}{Photo} & 
\multicolumn{2}{L}{Posing} & 
\multicolumn{2}{L}{Purchases} \\
Terms & short & long & short & long & short & long& short & long & short & long & short & long & short & long& short & long \\
\midrule
LSTM-3LR \cite{ERD}             & - & - & 2.25&2.93& 1.35& 3.42 & - & - & - & - & - & - & - & - & - & - \\ 
ERD \cite{ERD}                  & - & - &2.68& 2.92& 1.66 & 2.41& - & - & - & - & - & - & - & - & - & - \\
SRNN \cite{SRNN}                & - & - & 1.83& -& 1.35& - & - & - & - & - & - & - & - & - & - & - \\
Res-GRU \cite{Res-gru}          & 1.27 &1.59 & 1.45 & 1.86& 0.92 & 1.34 & 2.19 & 2.03& 0.88 & 1.89& 1.64 & 2.56 & 1.57&2.30& 1.51 & 2.31 \\
DAE-LSTM \cite{DAE}             & - & - & 1.53& 1.73& 1.86 & 2.01 & - & - & - & - & - & - & - & - & - & - \\
TE \cite{TE}                    & - & - & 0.18& 0.22 & 0.17 & 0.26& - & - & - & - & - & - & - & - & - & - \\ 
HP-GAN \cite{HP-gan}            & - & - & 0.91& 1.77 & 0.70 & 1.20 & - & - & - & - & - & - & - & - & - & - \\ 
C-seq2seq \cite{Cs2s}           & 0.80 & 1.45 & 0.94& 1.86 & 0.58& 1.24 & 1.21 & 1.72 & 1.51 & 1.81 & - & - & 1.12 & 2.65 & 1.19 & 2.52 \\ 
AGED \cite{AGED}                & 0.63 & - & 0.76& 1.30 & 0.51 & 0.93 & 1.30 & - & 0.50 & - & 0.81 & - & 1.12 & - & 1.01 & - \\
PAML \cite{PAML}                & - & - & -& - & - & - & - & - & - & - & - & - & - & - & - & - \\
MHU \cite{MHU}                  & - & - & 0.93 & 1.88 & - & - & 1.27 & 1.87 & - & 0.84 & 1.35 & 1.22 & 2.51 & - & - & - \\
QuaterNet \cite{QN}             & - & - & 0.85& - & 0.58 & - & - & - & - & - & - & - & - & - & - & - \\
BiHMP-GAN \cite{Bigan}          & - & - & 0.91& 1.77 & 0.54 & 1.20 & - & - & - & - & - & - & - & - & - & - \\
SkelNet \cite{SKEL}             & - & - & 0.90& - & 0.55 & - & - & - & - & - & - & - & - & - & - & - \\
CHA \cite{CHA}                  & 0.79 & - & 0.85& - & 0.53 & - & 1.28 & - & 1.51 & - & 0.80 &- & 1.16 & - & 1.15 & - \\
DSR \cite{DSR}                  & - & - & 0.91& - & 0.57 & - & - & - & - & - & - & - & - & - & - & - \\
VGRU \cite{VGRU}                & - & - & 0.95& - & 0.64 & - & - & - & - & - & - & - & - & - & - & - \\
HMR \cite{HMR}                  & - & - & 0.83& 1.72 & - & - & 1.25 & 1.73 & - & - & - & - & 1.12 & 2.50 & - & - \\
LTD \cite{LTD}                  & 0.71 & - & 0.77& - & 0.50 & - & 0.58 & - & 1.35 & 0.58 & 1.01 & - & 1.03 & - & 1.05 & - \\
RNN-SPL \cite{RNN-SPL}          & - & - & 0.95& - & 0.55 & - & - & - & - & - & - & - & - & - & - & - \\
GAN-poser \cite{Gan-poser}      & - & - & 1.11& - & 0.97 & - & - & - & - & - & - & - & - & - & - & - \\
AT-seq2seq \cite{ATs2s}         & - & - & 0.56& - & 0.49 & - & - & - & - & - & - & - & - & - & - & - \\
ARNet \cite{ARNET}              & 0.65 & - & 0.81 & - & 0.49 & - & 0.90 & - & 1.28 & - & 0.55 & - & 0.97 & - & 1.00 & - \\
Mix-and-Match \cite{mix}        & - & - & 0.83& 1.30 & 0.41 & 0.91 & - & - & - & - & - & - & - & - & - & - \\
MGCN \cite{MGCN}               & 0.65 & - & 0.92 & - & 0.49 & - & 0.94 & - & 1.29 & - & 0.58 & - & 1.06 & - & 1.05 & - \\
LDR \cite{LDR}                  & 0.59 & 0.95 & 0.72 & 0.84 & 0.49 & 0.97 & 0.87 & 1.33 & 0.63 & 1.33 & 0.45 & 1.20 & 0.91 & 1.34 \\
HRI \cite{HRI}                  & 0.69 & 1.27 & 0.87 & 1.63 & 0.60 & 1.10 & 0.95 & 1.57 & 1.31 & 1.68 & 0.58 & 1.08 & 1.09 & 2.32 & 1.00 & 2.22 \\
LPJP \cite{LPJP}                & - & - & 0.62 & - & 0.50 & - & 0.94 & - & - & - & - & - & - & - & - & - \\
MoPredNet \cite{MoP}            & - & - & 0.90 & 1.17 & 0.45 & 0.73 & - & - & - & - & - & - & - & - & - & - \\
MMA \cite{HR4}                  & 0.63 & 1.30 & 0.77 & 1.71 & 0.47 & 1.09 & 0.92 & 1.56 & 1.31 & 1.67 & 0.57 & 1.57 & 1.03 & 2.33 & 1.00 & 2.25  \\
AM-GAN \cite{AMGAN}             & 0.75 & 1.41 & 0.81 & 1.58 & 0.49 & 1.15 & 1.24 & 1.70 & 14.8 & 1.79 & 0.83 & 1.40 & 1.10 & 2.48 & 1.18 & 1.95 \\
NAT \cite{NAT}                  & 0.58 & 1.15 & 0.72 & 1.22 & 0.48 & 0.87 & 0.79 & 1.24 & 1.15 & 1.40 & 0.54 & 1.04 & 0.81 & 1.88 & 0.85 & 1.81 \\
Q-DCRN \cite{QDCRN}             & 0.62 & - & 0.98 & - & 0.56 & - & 1.11 & - & 1.39 & - & 0.64 & - & 1.28 & - & 1.08 & - \\
TCGAN \cite{TCGAN}              & 0.59 & - & 0.77 & - & 0.46 & - & 0.98 & - & 0.47 & - & 0.73 & - & 0.87 & - & 0.75 & - \\
yCNN \cite{Tracnn}     & - & - & -& - & - & - & - & - & - & - & - & - & - & - & - & - \\
DA-GAN \cite{DA-GAN}            & 0.56 & - & 0.77 & - & 0.49 & - & 0.87 & - & 0.63 & - & 0.51 & - & 0.94 & - & 0.92 & - \\
MT-GCN \cite{MTGCN}             & - & - & -& - & - & - & - & - & - & - & - & - & - & - & - & - \\
POTR \cite{POTR}                & 0.79 & - & 0.85 & - & 0.53 & - & 1.17 & - & 1.50 & - & 0.71 & - & 1.18 & - & 1.04 & - \\
LMC \cite{MGCN}                 & - & - & -& - & - & - & - & - & - & - & - & - & - & - & - & - \\
PVRED \cite{PVRED}              & 0.66 & - & 0.63 & - & 0.54 & - & 1.00 & - & 1.26 & - & 0.66 & - & 1.19 & - & 1.05 & - \\
MST-GNN \cite{MSGNN}            & 0.60 & - & 0.83 & - & 0.47 & - & 0.95 & - & 1.25 & - & 0.57 & - & 0.98 & - & 0.97 & - \\
SGAN \cite{SGRU}                & - & - & -& - & - & - & - & - & - & - & - & - & - & - & - & - \\
JDM \cite{Su}                   & - & - & -& - & - & - & - & - & - & - & - & - & - & - & - & - \\
\bottomrule
\end{tabular*}
\end{table*}

\section{Comparison of Different Methods}
In this section, the comparison of different methods. Human motion prediction is conventionally classified into short-term prediction and long-term prediction. Short-term prediction is less than 400 $ms$ and long-term prediction is between 400 $ms$ and 1,000 $ms$. On this basis, the comparison of performance is drawnamong all the methods for human motion prediction in these years. These models are: 
LSTM-3LR \cite{ERD}, 
ERD \cite{ERD}, 
Res-GRU \cite{Res-gru}, 
SRNN \cite{SRNN}, 
DAE-LSTM \cite{DAE}, 
TE \cite{TE},
HP-GAN \cite{HP-gan}, 
C-seq2seq \cite{Cs2s},
AGED \cite{AGED},
PAML \cite{PAML},
MHU \cite{MHU},
QuaterNet \cite{QN},
BiHMP-GAN \cite{Bigan},
SkelNet \cite{SKEL},
CHA \cite{CHA},
DSR \cite{DSR},
VGRU \cite{VGRU},
HMR \cite{HMR},
STMI-GAN \cite{STMI-GAN},
LTD \cite{LTD},
RNN-SPL \cite{RNN-SPL},
GAN-poser \cite{Gan-poser},
At-seq2seq \cite{ATs2s}, 
ARNet \cite{ARNET}, 
Mix-and-Match \cite{mix},
C-RNN \cite{C-RNN},
MGCN \cite{MGCN}, 
LDR \cite{LDR}, 
Dlow \cite{Dlow},
HRI \cite{HRI},
LPJP \cite{LPJP},
MoPredNet \cite{MoP},
MMA \cite{HR4},
AM-GAN \cite{AMGAN},
NAT \cite{NAT}, 
Q-DCRN \cite{QN},
TCGAN \cite{TCGAN},
TrajectoryCNN \cite{Tracnn},
DA-GAN \cite{DA-GAN},
MT-GCN \cite{MTGCN},
POTR \cite{POTR},
LMC \cite{MGCN},
PVRNN \cite{PVRED},
MST-GNN \cite{MSGNN},
SGAN \cite{SGRU},
JDM \cite{Su}. 

\textbf{Short-term and long-term human motion prediction} \quad
In order to provide a performance comparison for different human motion prediction algorithms, we select the most widely used benchmark dataset: H36M. 
For human motion prediction, it is always classified into short-term prediction (less than 400 $ms$) and long-term prediction (400 - 1,000 $ms$). For demonstration purposes, we show representative short-term prediction (320 $ms$) and long-term prediction (1,000 $ms$) separately for comparison. Then, we chose the two commonly used evaluation indicators as evaluation standards:  Mean Angle Error (MAE), Mean Per Joint Position Error (MPJPE). We compare all actions from the H36M.

As is shown in Tab \ref{h36mmae1} and Tab \ref{h36mmae2}, the performance is evaluated using MAE on H36M. All the actions in the datasets are listed. The results are derived from published papers.
In the Tab \ref{h36mmp1} and Tab \ref{h36mmp2}, the performance is evaluated using MPJPE on H36M. We also list all the actions in the datasets. The results are also derived from published papers.

From the distribution of the results, it is not difficult to note that MAE has been the dominant evaluation. MPJPE has been popular in the last year due to the widespread use of convolution methods. 
From the vertical quantitative comparison in each table,it is found that the methods in recent years possess tremendous progress.The methods for evaluating performance using MPJPE have become active in recent years. This indicates that researchers are demanding a more comprehensive look on algorithm performance. The success rate of the prediction method is also rising year by year. The success of these methods, especially the recent approaches \cite{MSGNN, Su}, proves that reasonable prior knowledge and kinematic constraints facilitate human motion prediction. 

From the horizontal quantitative comparison in each table, it can be seen that there remain many difficulties to conquer in terms of long-term prediction. In contrast, the error rate for long-term predictions is at least twice that of short-term predictions. Improving the accuracy of long-term prediction is a terribly challenging task. For actions, it is still difficult to predict actions that are not cyclical and involve a large scale of movements, e.g. purchases, discussions, etc.
In summary, the prediction of long-term, non-periodic and large movements remains a huge challenge. Innovative network structures coupled with valid prior knowledge are more helpful in accomplishing the task.

\begin{table*}
[width=2\linewidth,cols=17,pos=h]
\scriptsize
\centering
\caption{Performance evaluation (in MAE) on H36M dataset for both short-term and long-term human motion prediction.}\label{h36mmae2}
\begin{tabular*}{\tblwidth}{@{} LLLLLLLLLLLLLLLL@{} }
\toprule
Types & 
\multicolumn{2}{L}{Sitting} & 
\multicolumn{2}{L}{SittingDown}& 
\multicolumn{2}{L}{Smoking} & 
\multicolumn{2}{L}{Waiting} & 
\multicolumn{2}{L}{WalkDog}& 
\multicolumn{2}{L}{Walking}& 
\multicolumn{2}{L}{WalkTogether} \\
Terms & short & long & short & long & short & long& short & long & short & long & short & long & short & long\\
\midrule
LSTM-3LR \cite{ERD}     & - & - & -& - & 2.04 & 3.42 & - & - & - & - & 1.29 &2.20& - & - \\
ERD \cite{ERD}          & - & - & -& - & 2.35& 3.82& - & - & - & - & 1.59 & 2.38 & - & - \\
SRNN \cite{SRNN}        & - & - & -& - & 1.94& 3.23& - & - & - & - & 1.16 & 2.13 & - & - \\
Res-GRU \cite{Res-gru}  & 1.36 & 2.14 & 1.17 & 2.72 & 1.31 & 1.83 & 1.55 & 2.34 & 0.95 & 1.86 & 1.31 & 1.14 & 1.25& 1.42 \\
DAE-LSTM \cite{DAE}     & - & - & -& - & 1.15 & 1.77  & - & - & - & - & 1.39 & 1.39 & - & - \\
TE \cite{TE}            & - & - & -& - & 0.19 & 0.27 & - & - & - & - & 0.20 & 0.24& - & - \\
HP-GAN \cite{HP-gan}    & - & - & -& - & 0.91 & 1.11 & - & - & - & - & - & 0.63 & 0.85 & - & - \\ 
C-seq2seq \cite{Cs2s}   & 1.02 & 1.67 & 1.16& 2.06 & 0.96 & 1.62 & 1.09 & 2.50 & 1.32 & 1.92 & 0.68 & 0.92 & 0.71 & 1.28 \\ 
AGED \cite{AGED}        & 1.05 & - & 0.98& - & 0.82 & 1.21 & 0.55 & - & 1.15 & - & 0.55 & 0.91 & 0.56 & -\\
PAML \cite{PAML}        & - & - & -& - & - & - & - & - & - & - & - & - & - & - \\
MHU \cite{MHU}          & - & - & -& - & - & - & - & - & 1.21 & 1.90 & 0.69 & 1.06 & - & - \\
QuaterNet \cite{QN}     & - & - & -& - & 0.93 & - & - & - & - & - & 0.56 & - & - & - \\
BiHMP-GAN \cite{Bigan}  & - & - & -& - & 0.91 & 1.11 & - & - & - & - & 0.67 & 0.85 & - & - \\
SkelNet \cite{SKEL}     & - & - & -& - & 0.91 & - & - & - & - & - & - & - & - & - \\
CHA \cite{CHA}          & 1.03 & - & 1.15& - & 0.89 & - & 1.12 & - & 0.75 & - & 1.16 & - & 0.75 & - \\
DSR \cite{DSR}          & - & - & -& - & 0.90 & - & - & - & - & - & 0.69 & - & - & - \\
VGRU \cite{VGRU}        & - & - & -& - & 0.85 & - & - & - & - & - & 0.64 & - & - & - \\
HMR \cite{HMR}          & - & - & -& - & - & - & - & - & 1.20 & 1.84 & - & - & - & - \\
LTD \cite{LTD}          & -0.80 & - & 0.90& - & 0.86 & - & 0.91 & - & 1.12 & - & 0.49 & - & 0.52 & - \\
RNN-SPL \cite{RNN-SPL}  & - & - & -& - & 0.96 & - & - & - & - & - & 0.67 & - & - & - \\
GAN-poser \cite{Gan-poser} & - & - & -& - & 0.87 & - & - & - & - & - & 0.84 & - & - & - \\
AT-seq2seq \cite{ATs2s} & - & - & -& - & 0.49 & - & - & - & - & - & 0.51 & - & - & - \\
ARNet \cite{ARNET}      & 0.80 & - & 0.87 & - & 0.86 & - & 0.90 & - & 1.11 & - & 0.49 & - & 0.53& - \\
Mix-and-Match \cite{mix}& - & - & -& - & 0.79 & 1.25 & - & - & - & - & 0.56& 0.68 & - & - \\
MGCN \cite{MGCN}       & 0.76 & - & 0.93 & - & 0.81 & - & 0.88 & - & 1.16 & - & 0.49 & - & 0.50 & - \\
LDR \cite{LDR}          & 0.69 & 1.38 & 0.87 & 1.42 & 0.79 & 1.08 & 0.84 & 1.21 & 0.93 & 1.38 & 0.46 & 0.71 & 0.49 & 1.38 \\
HRI \cite{HRI}          & 0.83 & 1.55 & 0.92 & 1.70 & 0.86 & 2.30 & 0.92 & 1.82 & 1.05 & 0.64 & 0.46 & 1.16 & 1.07 & 2.22 \\
LPJP \cite{LPJP}        & - & - & -& - & 0.85 & - & - & - & - & - & 0.51 & - & - & - \\
MoPredNet \cite{MoP}    & - & - & -& - & 0.53 & 1.88 & - & - & - & - & 0.43 & 0.83 & - & - \\
MMA \cite{HR4}          & 0.81 & 1.54 & 0.91 & 1.68 & 0.86 & 1.57 & 0.90 & 2.27 & 1.05 & 1.81 & 0.48 & 0.63 & 0.50 & 1.18 \\
AM-GAN \cite{AMGAN}     & 0.98 & 1.60 & 1.08 & 1.85 & 0.88 & 1.10 & 1.10 & 2.15 & 1.18 & 1.80 & 0.62 & 0.84 & 0.71 & 1.19 \\
NAT \cite{NAT}          & 0.84 & 1.46 & 0.94 & 1.58 & 0.81 & 1.26  & 0.79 & 1.58 & 0.86 & 1.44 & 0.45 & 0.50 & 0.51 & 1.07 \\
Q-DCRN \cite{QDCRN}     & 0.88 & - & 1.03 & - & 0.87 & - & 0.99 & - & 1.10 & - & 0.56 & - & 0.57 & - \\
TCGAN \cite{TCGAN}      & 0.79 & - & 0.74 & - & 0.67 & - & 0.74 & - & 1.01 & - & 0.52 & - & 0.51 & -  \\
TrajectoryCNN \cite{Tracnn} & - & - & -& - & - & - & - & - & - & - & - & - & - & - \\
DA-GAN \cite{DA-GAN}    & 0.73 & - & 0.86 & - & 0.75 & - & 0.85 & - & 1.02 & - & 0.44 & - & 0.46 & - \\
MT-GCN \cite{MTGCN}     & - & - & -& - & - & - & - & - & - & - & - & - & - & - \\
POTR \cite{POTR}        & 0.92 & - & 1.00 & - & 0.84 & - & 1.14 & - & 1.21 & - & 0.62 & - & 0.63 & - \\
LMC \cite{MGCN}         & - & - & -& - & - & - & - & - & - & - & - & - & - & - \\
PVRED \cite{PVRED}      & 0.84 & - & 1.03 & - & 0.81 & - & 0.93 & - & 1.13 & - & 0.54 & - & 0.59 & - \\
MST-GNN \cite{MSGNN}    & 0.75 & - & 0.88 & - & 0.78 & - 0.88 & - & 1.11 & - & 0.49 & - & 0.51 & -  \\
SGAN \cite{SGRU}        & - & - & -& - & - & - & - & - & - & - & - & - & - & - \\
JDM \cite{Su}           & - & - & -& - & - & - & - & - & - & - & - & - & - & - \\
\bottomrule
\end{tabular*}
\end{table*}
\begin{table*}
[width=2\linewidth,cols=17,pos=h]
\renewcommand\arraystretch{1.1} 
\scriptsize
\centering
\caption{Performance evaluation (in MPJPE) on H36M dataset for both short-term and long-term human motion prediction.}\label{h36mmp1}
\begin{tabular*}{\tblwidth}{@{} LLLLLLLLLLLLLLLLL @{} }
\toprule
Types & 
\multicolumn{2}{L}{Directions}& 
\multicolumn{2}{L}{Discussion} & 
\multicolumn{2}{L}{Eating} & 
\multicolumn{2}{L}{Greeting}& 
\multicolumn{2}{L}{Phoning}& 
\multicolumn{2}{L}{Photo} & 
\multicolumn{2}{L}{Posing} & 
\multicolumn{2}{L}{Purchases} \\
Terms & short & long & short & long & short & long& short & long & short & long & short & long & short & long& short & long \\
\midrule
LSTM-3LR \cite{ERD}             & 46.6 & 135.1 & 85.5 & 135.1 & 78.1 & 121.1 & 73.1 & 177.6 & 68.8 & 133.3 & 71.1 & 155.6 & 130.1 & 176.5 & 88.0 & 142.9 \\ 
ERD \cite{ERD}                  & - & - & -& - & - & - & - & - & - & - & - & - & - & - & - & - \\
SRNN \cite{SRNN}                & - & - & -& - & - & - & - & - & - & - & - & - & - & - & - & - \\
Res-GRU \cite{Res-gru}          & - & - & -& - & - & - & - & - & - & - & - & - & - & - & - & - \\
DAE-LSTM \cite{DAE}             & - & - & -& - & - & - & - & - & - & - & - & - & - & - & - & - \\
TE \cite{TE}                    & - & - & -& - & - & - & - & - & - & - & - & - & - & - & - & - \\ 
HP-GAN \cite{HP-gan}            & - & - & -& - & - & - & - & - & - & - & - & - & - & - & - & - \\ 
C-seq2seq \cite{Cs2s}           & 44.5 & 78.3 & 35.2 & 67.4 & 37.2 & 53.8 & 66.3 & 129.7 & 29.6 & 116.4 & 33.1 & 85.8 & 58.3 & 113.7 & 62.8 & 112.6 \\ 
AGED \cite{AGED}                & - & - & -& - & - & - & - & - & - & - & - & - & - & - & - & - \\
PAML \cite{PAML}                & - & - & -& - & - & - & - & - & - & - & - & - & - & - & - & - \\
MHU \cite{MHU}                  & - & - & -& - & - & - & - & - & - & - & - & - & - & - & - & - \\
QuaterNet \cite{QN}             & - & - & -& - & - & - & - & - & - & - & - & - & - & - & - & - \\
BiHMP-GAN \cite{Bigan}          & - & - & -& - & - & - & - & - & - & - & - & - & - & - & - & - \\
SkelNet \cite{SKEL}             & - & - & -& - & - & - & - & - & - & - & - & - & - & - & - & - \\
CHA \cite{CHA}                  & - & - & -& - & - & - & - & - & - & - & - & - & - & - & - & - \\
DSR \cite{DSR}                  & - & - & -& - & - & - & - & - & - & - & - & - & - & - & - & - \\
VGRU \cite{VGRU}                & - & - & -& - & - & - & - & - & - & - & - & - & - & - & - & - \\
HMR \cite{HMR}                  & - & - & -& - & - & - & - & - & - & - & - & - & - & - & - & - \\
LTD \cite{LTD}                  & 48.2 & 89.1 & 39.6 & 78.5 &25.3 & 44.3& 74.2 & 148.4 & 37.9 & 94.3 & 38.2 & 125.7 & 66.2 & 143.5 & 64.4 & 127.2  \\
RNN-SPL \cite{RNN-SPL}          & - & - & -& - & - & - & - & - & - & - & - & - & - & - & - & - \\
GAN-poser \cite{Gan-poser}      & - & - & -& - & - & - & - & - & - & - & - & - & - & - & - & - \\
AT-seq2seq \cite{ATs2s}         & - & - & -& - & - & - & - & - & - & - & - & - & - & - & - & - \\
ARNet \cite{ARNET}              & - & - & -& - & - & - & - & - & - & - & - & - & - & - & - & - \\
Mix-and-Match \cite{mix}        & - & - & -& - & - & - & - & - & - & - & - & - & - & - & - & - \\
MGCN \cite{MGCN}               & - & - & -& - & - & - & - & - & - & - & - & - & - & - & - & - \\
LDR \cite{LDR}                  & 47.2 & 106.5 & 46.3 & 144.6 & 24.8 & 43.1 & 47.2 & 127.3 & 87.9 & 143.2 & 25.4 & 45.8 & 31.2 & 112.6 \\
HRI \cite{HRI}                  & 44.5 & 106.5 & 52.1 & 119.8 & 28.7 & 75.7 & 63.8 & 105.0 & 39.0 & 115.9 & 40.7 & 178.2 & 58.5 & 138.8 & 60.4 & 134.2 \\
LPJP \cite{LPJP}                & - & - & 48.0 & - & 37.4 & - & 64.5 & - & - & - & - & - & - & - & - & - \\
MoPredNet \cite{MoP}            & - & - & -& - & - & - & - & - & - & - & - & - & - & - & - & - \\
MMA \cite{HR4}                  & 50.6 & 105.7 & 37.7 & 117.5 & 45.2 & 73.7 & 68.2 & 130.7 & 37.7 & 104.6 & 38.0 & 115.2 & 62.2 & 172.9 & 58.4 & 115.0 \\
AM-GAN \cite{AMGAN}             & - & - & -& - & - & - & - & - & - & - & - & - & - & - & - & - \\
NAT \cite{NAT}                  & 30.7 & - & 41.5 & - & 33.8 & - & 46.0 & - & 35.5 & - & 35.0 & - & 50.7 & - & 50.1 & - \\
Q-DCRN \cite{QDCRN}             & - & - & -& - & - & - & - & - & - & - & - & - & - & - & - & - \\
TCGAN \cite{TCGAN}              & - & - & -& - & - & - & - & - & - & - & - & - & - & - & - & - \\
TrajectoryCNN \cite{Tracnn}     & 50.2 & 104.2 & 41.3 & 103.0 & 37.0 & 71.5 & 67.3 & 84.3 & 37.0 & 113.5 & 36.2 & 86.6 & 62.9 & 210.9 & 64.3 & 115.5 \\
DA-GAN \cite{DA-GAN}            & - & - & -& - & - & - & - & - & - & - & - & - & - & - & - & - \\
MT-GCN \cite{MTGCN}             & 85.3 & 147.9 & 70.2 & 131.2 & 53.0 & 89.2 & 77.8 & 151.2 & 44.2 & 139.9 & 49.6 & 137.1 & 70.3 & 176.4 & 74.2 & 149.3 \\
POTR \cite{POTR}                & - & - & -& - & - & - & - & - & - & - & - & - & - & - & - & - \\
LMC \cite{MGCN}                 & - & - & 48.9 & - & 37.7 & - & 68.4 & - & - & - & - & - & - & - & - & - \\
PVRED \cite{PVRED}              & - & - & -& - & - & - & - & - & - & - & - & - & - & - & - & - \\
MST-GNN \cite{MSGNN}            & - & - & -& - & - & - & - & - & - & - & - & - & - & - & - & - \\
SGAN \cite{SGRU}                & 42.5 & 85.8 & 31.1 & 59.8 & 24.0 & 41.5 & 40.0 & 92.1 & 28.9 & 80.6 & 25.2 & 85.5 & 29.9 & 80.6 & 51.2 & 102.2 \\
JDM \cite{Su}                   & 37.6 & 72.3 & - & - & 21.0 & 52.5 & 46.8 & 83.1 & 24.3 & 102.7 & 23.8 & 82.4 & 49.2 & 123.8 & 52.3 & 112.2  \\
\bottomrule
\end{tabular*}
\end{table*}

\begin{table*}
[width=2\linewidth,cols=17,pos=h]
\renewcommand\arraystretch{1} 
\footnotesize
\centering
\caption{Performance evaluation (in MPJPE) on H36M dataset for both short-term and long-term human motion prediction.}\label{h36mmp2}
\begin{tabular*}{\tblwidth}{@{} LLLLLLLLLLLLLLLLL @{} }
\toprule
Types & 
\multicolumn{2}{L}{Sitting} & 
\multicolumn{2}{L}{SittingDown}& 
\multicolumn{2}{L}{Smoking} & 
\multicolumn{2}{L}{Waiting} & 
\multicolumn{2}{L}{WalkDog}& 
\multicolumn{2}{L}{Walking}& 
\multicolumn{2}{L}{WalkTogether} \\
Terms & short & long & short & long & short & long& short & long & short & long & short & long & short & long \\
\midrule
LSTM-3LR \cite{ERD}             & - & - & -& - & - & - & - & - & - & - & - & - & - & -  \\ 
ERD \cite{ERD}                  & - & - & -& - & - & - & - & - & - & - & - & - & - & -  \\
SRNN \cite{SRNN}                & - & - & -& - & - & - & - & - & - & - & - & - & - & -  \\
Res-GRU \cite{Res-gru}          & - & - & -& - & - & - & - & - & - & - & - & - & - & -  \\
DAE-LSTM \cite{DAE}             & - & - & -& - & - & - & - & - & - & - & - & - & - & -  \\
TE \cite{TE}                    & - & - & -& - & - & - & - & - & - & - & - & - & - & -  \\ 
HP-GAN \cite{HP-gan}            & - & - & -& - & - & - & - & - & - & - & - & - & - & -  \\ 
C-seq2seq \cite{Cs2s}           & 47.2 & 106.5 & 46.3 & 144.6 & 24.8 & 43.1 & 47.2 & 127.3 & 87.9 & 143.2 & 25.4 & 45.8 & 31.2 & 79.2\\ 
AGED \cite{AGED}                & - & - & -& - & - & - & - & - & - & - & - & - & - & -  \\
PAML \cite{PAML}                & - & - & -& - & - & - & - & - & - & - & - & - & - & - \\
MHU \cite{MHU}                  & - & - & -& - & - & - & - & - & - & - & - & - & - & - \\
QuaterNet \cite{QN}             & - & - & -& - & - & - & - & - & - & - & - & - & - & -  \\
BiHMP-GAN \cite{Bigan}          & - & - & -& - & - & - & - & - & - & - & - & - & - & - \\
SkelNet \cite{SKEL}             & - & - & -& - & - & - & - & - & - & - & - & - & - & - \\
CHA \cite{CHA}                  & - & - & -& - & - & - & - & - & - & - & - & - & - & -  \\
DSR \cite{DSR}                  & - & - & -& - & - & - & - & - & - & - & - & - & - & -  \\
VGRU \cite{VGRU}                & - & - & -& - & - & - & - & - & - & - & - & - & - & -  \\
HMR \cite{HMR}                  & - & - & -& - & - & - & - & - & - & - & - & - & - & -  \\
LTD \cite{LTD}                  & 24.6 & 119.8 & 56.4 & 163.9 & 25.3 & 44.3 & 57.5 & 157.2 & 102.2 & 185.4 & 29.2 & 50.9 & 35.3 & 102.4 \\
RNN-SPL \cite{RNN-SPL}          & - & - & -& - & - & - & - & - & - & - & - & - & - & -  \\
GAN-poser \cite{Gan-poser}      & - & - & -& - & - & - & - & - & - & - & - & - & - & -  \\
AT-seq2seq \cite{ATs2s}         & - & - & -& - & - & - & - & - & - & - & - & - & - & -  \\
ARNet \cite{ARNET}              & - & - & -& - & - & - & - & - & - & - & - & - & - & -  \\
Mix-and-Match \cite{mix}        & - & - & -& - & - & - & - & - & - & - & - & - & - & -  \\
MGCN \cite{MGCN}               & - & - & -& - & - & - & - & - & - & - & - & - & - & -  \\
LDR \cite{LDR}                  & 44.5 & 78.3 & 35.2 & 67.4 & 37.2 & 53.8 & 66.3 & 129.7 & 33.1 & 85.8 & 29.6 & 116.4 & 58.3 & 133.7 \\
HRI \cite{HRI}                  & 44.3 & 115.9 & 59.1 & 143.6 & 29.9 & 69.5 & 43.4 & 108.2 & 73.3 & 146.9 & 34.2 & 58.1 & 35.1 & 64.9\\
LPJP \cite{LPJP}                & - & - & -& - & 24.1 & - & - & - & - & 29.1 & - & - & - & -  \\
MoPredNet \cite{MoP}            & - & - & -& - & - & - & - & - & - & - & - & - & - & - \\
MMA \cite{HR4}                  & 53.8 & 115.0 & 54.6 & 141.8 & 25.4 & 68.7& 55.7 & 105.1 & 100.3 & 141.4 & 25.5 & 57.1 & 33.2 & 63.2 \\
AM-GAN \cite{AMGAN}             & - & - & -& - & - & - & - & - & - & - & - & - & - & - \\
NAT \cite{NAT}                  & 44.0 & - & 34.8 & - & 23.8 & - & 50.5 & - & 55.3 & - & 23.3 & - & 28.0 & -  \\
Q-DCRN \cite{QDCRN}             & - & - & -& - & - & - & - & - & - & - & - & - & - & -  \\
TCGAN \cite{TCGAN}              & - & - & -& - & - & - & - & - & - & - & - & - & - & - \\
TrajectoryCNN \cite{Tracnn}     & 49.4 & 116.3 & 55.1 & 123.8 & 23.7 & 58.7 & 53.4 & 165.9 & 98.1 & 181.3 & 30.0 & 46.4 & 33.9 & 77.3  \\
DA-GAN \cite{DA-GAN}            & - & - & -& - & - & - & - & - & - & - & - & - & - & -  \\
MT-GCN \cite{MTGCN}             & 56.7 & 131.8 & 60.2 & 158.5 & 43.9 & 85.5 & 57.7 & 130.9 & 92.5 & 176.3 & 45.6 & 78.3 & 43.1 & 80.2  \\
POTR \cite{POTR}                & - & - & -& - & - & - & - & - & - & - & - & - & - & -  \\
LMC \cite{MGCN}                 & - & - & -& - & 22.6 & - & - & - & - & - & 27.1 & - & - & -  \\
PVRED \cite{PVRED}              & - & - & -& - & - & - & - & - & - & - & - & - & - & -  \\
MST-GNN \cite{MSGNN}            & - & - & -& - & - & - & - & - & - & - & - & - & - & -  \\
SGAN \cite{SGRU}                & 40.2 & 99.8 & 38.2 & 113.2 & 40.0 & 92.1 & 69.3 & 131.5 & 22.6 & 39.8 & - & - & 28.8 & 75.2  \\
JDM \cite{Su}                   & 36.9 & 93.6 & 32.7& 101.9 & - & - & 37.5 & 95.6 & 62.3 & 126.1 & - & - & - & -  \\
\bottomrule
\end{tabular*}
\end{table*}

\section{Discussion}

In the past few years, striking progress has been made in the development of advanced prediction technology in terms of method diversity, performance, and correlation application scenarios. More issues worthy of our discussion have also emerged. In this section, two questions for the 3D human motion prediction are discussed. 

\textbf{Q1}: Are evaluation criteria effective to measure performance? This will be discussed in Sec. 6.1 by reviewing the existing benchmarking practices including metrics, experiments, and datasets.

\textbf{Q2}: How should relevant research be further carried out in the future? This will be discussed in Sec. 6.2 by outlining open challenges and potential future research directions.

\subsection{Benchmarking}

Evaluating the performance of motion prediction algorithms requires the selection of appropriate experimental scenarios and accuracy metrics, as well as the robustness of research methods and a large number of diverse datasets.

    Existing datasets, summarized in sec.4.1, are varied and contain a wealth of information. However, datasets are missed when regarding the interaction between humans and the environment. The environment is a highly decisive factor to human motion. Further, all of the motion scenes are solitary, which brings some limitations for researching multiple persons' motions. Furthermore, existing datasets tend to be more deterministic prediction than probabilistic prediction. 

As is shown in sec.4.2, the measurement of human motion prediction has revealed a problem that needs to be improved. \textit{For the accuracy metrics}, there is a lack of specific measurements for long-term prediction. Only a measurement strategy to this problem is presented in the method. Now that we've been trying to divide forecasts into long-term and short-term, it is a bit inappropriate to merely use the same metric.
\textit{For the probabilistic metrics}, it reflects the stochastic nature of human motion. Although more and more colleagues have been engaging in research in this area, there are no effective metrics for probabilistic prediction. Moreover, current methods rely more on the comparison of visual results.  

\subsection{Future Directions}

To solve complex problems with simple methods is the pursuit of all research. Vague understandings of the problems intensify the intricacy of problems.
Nowadays, more and more advanced techniques in different fields are being applied to human motion prediction. Furthermore, some methods carry a novel understanding of human motion, which promotes human motion prediction. 
With these trends in mind, we want to propose several potential and interesting directions for future research. 

Prediction methods are driven by the intrinsic logic from researchers’ understanding of human motion.
As is shown in Fig.\ref{FIG:class}, Our proposed directions are inseparable from the pyramid representation of human motion prediction (\textit{Human pose representation}, \textit{Network structure design}, and \textit{Prediction target}). 

For example, the most efficient RNNs structure is utilized to predict future poses, because human motion prediction is a sequential seq2seq task. 
Firstly, the form of data (sequential data) is differentiated. Then, a pertinent model will be selected to deal with the data. Finally, the output will in return test our method.
At the same time, it becomes known to us that human motion is different from the general seq2seq task because it is subject to kinematic constraints. 
Therefore, many kinematic methods were developed, such as Graph modeling, kinematics tree modeling to represent the human pose. Further, the new raised mathematical thinking that can map the human body into other spaces will become another solid approach. 
Furthermore, more can be learnt from other related disciplines, which benefits our understanding of the underlying logic of the problem. Such as, quantum physics, mechanics, anatomy, kinematics, causal intervention, etc.
Owning to these logics, the human pose representations will become more diverse and multi-scaled. Correspondingly, matched network is naturally sought or designed to learn about them.  Likewise, our prediction target also influences the network. In summary, the above perspectives of thinking inspires a three-phase idea for the future directions: \textit{Data phase}, \textit{Network phase}, and  \textit{Prediction result phase}. 

\textbf{Data phase}\quad 
In the data phase, there are mostly two directions: \textit{richer datasets} and \textit{effective pose representation}. 
\textit{For the former}, the current datasets are only captured single pose data. However, these datasets disregard the external stimuli (physical environment, passers-by) in the real world. This may cause limitations to the application of the methods using these datasets, such as the influence of interaction between human and environment (or others) on motion prediction. Moreover, existing datasets are all targeted at the deterministic prediction, which are inconsistent with objective facts.
Therefore, a chief direction is to enrich the datasets, In which three means can be considered: 
($i$) Adding interactive physical environment elements, such as desk, chair, and door. 
($ii$) Capturing multiple poses simultaneously. 
($iii$) Establishing probabilistic motion prediction datasets.
\textit{For latter}, Our mindsets should not be confined to apparent ways, like kinematics and anatomy, but should be expanded to different fields, such as advanced mathematics, quantum mechanics, and so on.

\textbf{Network phase}\quad
The network is vital for any deep learning work. The objective is to extract and process powerful features for the task. Yet, the human motion prediction task possesses a prior kinematic constraint.  Most of the presented methods for human motion prediction are designed for specific tasks or scenarios. These make the network less robust. Practically speaking, the applicability of the network will thus be hindered. Therefore, a robust network needs to be proposed, so that it can adapt to different actions or situations. Meanwhile, interpretability is always associated with deep learning. The role of each particular layer cannot be precisely determined. Therefore, there is a lack of systematic guidance when designing networks.
Besides, current methods mainly focus on the model structure, failing to notice the importance of network training. An effective training strategy can improve network performance. To sum up, there are three ways for improving the network performance:
($i$) A robust network.
($ii$) An interpretable network.
($iii$) An efficient network training method.

\textbf{Prediction result phase}\quad
In the prediction result phase, there are mainly two directions: \textit{Prediction target} and \textit{Evaluation criteria}. 
For the \textit{former}, 
human motion prediction is a kind of motion generation, whether probabilistic prediction or deterministic prediction. Beyond this, multiple targets can be produced along with this idea. ($i$) Action transformation. One action can smoothly transfer to another one. 
($ii$) Longer-term motion prediction. For existing methods, their long-term prediction is from 400 $ms$ to 1,000 $ms$. Most of the methods tend to degrade into motionless states or drift away to non-human-like motions for longer-term prediction. Thereby, it is of extraordinary research value.
And probabilistic prediction or deterministic prediction is surely worth studying as well.
For the \textit{latter}, the existing evaluation criteria mainly evaluate the deterministic prediction. For other means (e.g., probabilistic prediction and action transfer), the researchers only depend on visual results. However, This can only result in the making of an intuitive judgment on the results with a large gap, which is rather unfavorable to the research. Therefore, it is imperative to propose a new evaluation criterion for more diverse researches.

\section{Conclusion}

A survey of the 3D human motion prediction methods is presented in this paper. The literatures across multiple domains are involved in the survey, and a taxonomy of human motion prediction approaches is presented. The existing 3D human motion prediction tasks are divided into three general categories: \textit{Human pose representation}, \textit{Network structure design}, and \textit{Prediction target}. Accordingly, the existing methods are combed in detail. Moreover, the metrics of evaluating these methods are summarized and extensive experimental results are illustrated. 
In the end, we retrospect and discussed the existing methods and proposed three potential future research directions.  
We hope this survey can contribute to the progress of the field.  

\section{Acknowledgement}
This research is partly supported by National Key Research and Development Program of China, Grant Number: 2018YFB0804202, 2018YFB0804203; Regional Joint Fund of NSFC, Grant Number: U19A2057; National Natural Science Foundation of China, Grant Number: 61876070; Jilin Province Science and Technology Development Plan Project, Grant Number: 20190303134SF.
\bibliographystyle{cas-model2-names}
\bibliography{Survey}

\bio{A}
Kedi Lyu received the B.E. and B.S. degrees from the Jilin University, Changchun, China, in 2016 and 2019, respectively. He is currently pursuing the Ph.D. degree with Jilin University, Changchun, China. His research interests include computer vision, machine learning, and deep learning.
\endbio

\bio{B}
Haipeng Chen is a professor of Jilin University, Changchun, China. He received his Ph.D. and B.E. degrees from Jilin University, Changchun, China. His research interests include computer vision, machine learning.
\endbio

\bio{C}
Zhenguang Liu is a professor of Zhejiang Gongshang University, Hangzhou, China. He had been a research fellow in National University of Singapore and A*STAR (Agency for Science, Technology and Research, Singapore) for several years. He respectively received his Ph.D. and B.E. degrees from Zhejiang University and Shandong University, China. His research interests include multimedia data analysis and smart contract security. Various parts of his work have been published in top-tier venues including TIP, CVPR, ICCV, TKDE, AAAI, ACM MM, INFOCOM, IJCAI, WWW, TMC, WWW. Dr. Liu has served as technical program committee member for top-tier conferences such as ACM MM, CVPR, AAAI, IJCAI, and ICCV, session chair of ICGIP, local chair of KSEM, and reviewer for top-tier journals IEEE TVCG, IEEE TPDS, IEEE TMM, ACM TOMM.
\endbio

\bio{D}
Beiqi Zhang received his bachelor's degree from Sichuan University in 2020. His current research interest is multimedia data analysis. Recently he has been researching on human motion and pose estimation at Zhejiang Gongshang University. 

\endbio

\bio{E}
Ruili Wang is a Professor in the School of Computer and Information Engineering at Zhejiang Gongshang University. His research areas include image and video processing, speech and natural language processing, machine learning and deep learning, data mining and big data. He has supervised more than 20 PhD students to completion, and has published over 150 papers. He is an Associate Editor (or Editor Board member) for the journals of IEEE Transactions on Emerging Topics in Computational Intelligence, Neurocomputing (Elsevier), Applied Soft Computing (Elsevier), Knowledge and Information Systems (Springer), Health Information Science and Systems (Springer), and Complex and Intelligent Systems (Springer).
\endbio

\end{document}